 \documentclass[pmlr,twocolumn,10pt]{jmlr} 

\usepackage{booktabs}
\usepackage{siunitx}

\usepackage{float}

\theorembodyfont{\upshape}
\theoremheaderfont{\scshape}
\theorempostheader{:}
\theoremsep{\newline}

\jmlrvolume{LEAVE UNSET}
\jmlryear{2023}
\jmlrsubmitted{LEAVE UNSET}
\jmlrpublished{LEAVE UNSET}
\jmlrworkshop{Machine Learning for Health (ML4H) 2023} 

 \newcommand{\bs}[1]{\boldsymbol{#1}}        
 \newcommand{\RR}[1]{\mathbb{R}^{#1}}
   
    \DeclareMathOperator*{\argmax}{argmax}

\title[Generating Personalized Treatments]{Generating
Personalized 
Insulin 
Treatments
Strategies
with 
\\
Deep
Conditional 
  Generative Time Series
Models }

\author{%
\Name{Manuel Sch\"urch} \Email{manuel.schuerch@uzh.ch}\\
\addr University of Zurich, Switzerland
\AND
\Name{Xiang Li} \Email{xiang.li@outlook.de}\\
\addr University of Zurich, Switzerland
\AND
\Name{Ahmed Allam} \Email{ahmed.allam@uzh.ch}\\
\addr University of Zurich, Switzerland
\AND
\Name{Giulia Rathmes} \Email{giulia.rathmes@usz.ch}\\
\addr University Hospital of Zurich, Switzerland
\AND
\Name{Amina Mollaysa} \Email{maolaaisha.aminanmu@uzh.ch}\\
\addr University of Zurich, Switzerland
\AND
\Name{Claudia Cavelti-Weder} \Email{claudia.cavelti-weder@usz.ch}\\
\addr University Hospital of Zurich, Switzerland
\AND
\Name{Michael Krauthammer} \Email{michael.krauthammer@uzh.ch}\\
\addr University of Zurich, Switzerland
\AND
}

\begin{document}

\maketitle

\begin{abstract}
We propose a novel framework 
that combines deep generative time series models with decision theory
for generating 
personalized treatment strategies.
It leverages historical patient trajectory data to jointly learn the generation of realistic personalized 
treatment 
and 
future outcome trajectories through deep generative time series models. 
In particular, our framework enables the 
generation of novel multivariate treatment strategies tailored to the personalized patient history and trained for optimal expected future outcomes based on conditional expected utility maximization.
We demonstrate our framework by 
      generating personalized
     insulin treatment strategies and blood glucose predictions for hospitalized diabetes patients, showcasing the potential of our approach for generating improved personalized treatment strategies.
\end{abstract}

\section{Introduction}
\label{sec:intro}

Recent advancements in deep conditional generative models have revolutionized various domains including 
the generation of
textual prompt-answers 
\citep{brown2020language}, 
 high-quality images \citep{rombach2022high},
and molecules based on desired properties \citep{mollaysa2019conditional}. However, applying these techniques for generating complex and multivariate healthcare time series
poses unique challenges that have yet to be fully addressed. 
A fundamental issue is the lack of a clear and simple framework for leveraging conditional samples from multivariate time series.
%
%
%
%
%
%
%
 To tackle this, we explore the potential of combining deep probabilistic generative time series models \citep{Tomczak2022DeepModeling} with decision theory, 
 \begin{figure*}[htp!]
\centering
\floatconts
{fig:overview_generation_trajectories}
{\caption{\small Overview of data (left), our approach (middle) and objective (right)
for generating personalized treatments.
}}
{\includegraphics[width=1.\linewidth]{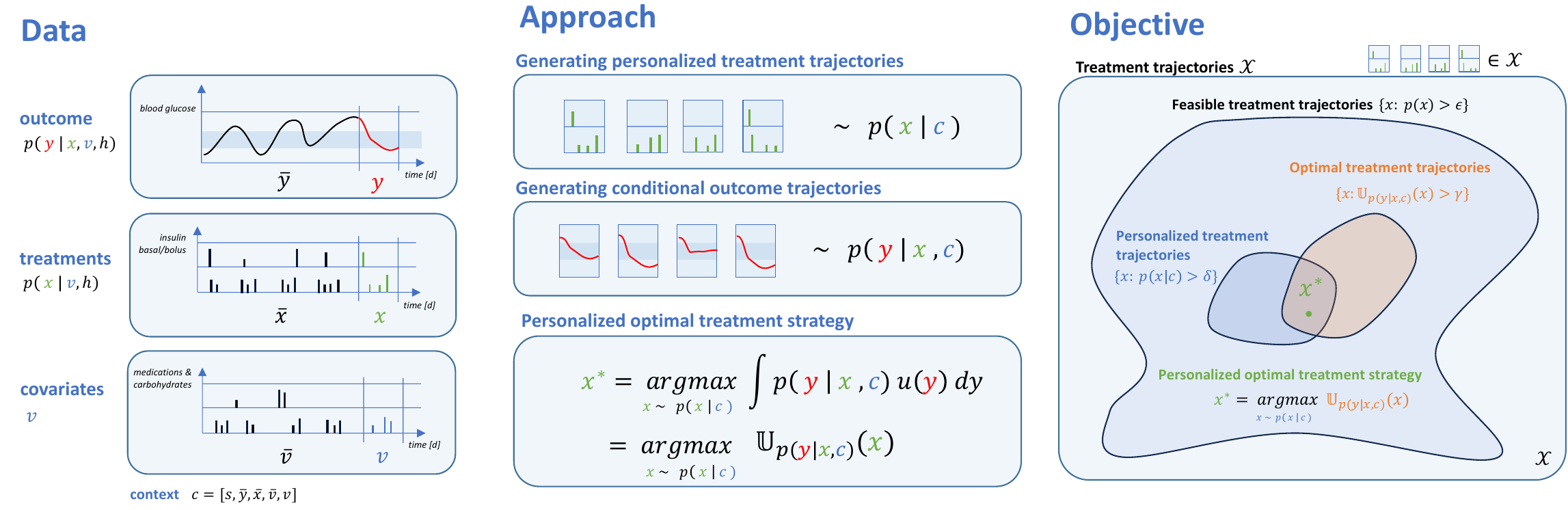}}
\end{figure*}
%
 to enable optimal personalized treatment generation learned from retrospective patient data.
By exploiting the power of conditional generative models, we can first learn to generate feasible  personalized treatment strategies
conditioned on the current personalized conditions. 
Second, we can also learn to generate future personalized outcome trajectories conditioned on specific treatment trajectories.
Third, to bridge the gap between the conditional generation of time series and decision-making, we propose 
 the 
 personalized
 treatment generation by optimizing 
 the 
 expected utility of the future outcome trajectories,
 so that both feasible and optimal treatments can be learned.
%
 \newline
%
%
%
%
%
%
Our proposed approach is illustrated in \autoref{fig:overview_generation_trajectories}
and 
offers several innovative contributions:
\begin{itemize}
\item 
\textbf{Framework for 
Time Series
Samples:
} 
We present 
a new framework  for leveraging 
samples of 
conditional
distributions over 
multivariate time series
based on 
decision theory.
\item 
\textbf{Generating 
Realistic
Treatments:
} 
We exploit deep generative models
 for generating 
 novel personalized 
 treatment 
 strategies 
 and
 future 
 outcome trajectories.
\item \textbf{Optimal Personalized Treatments:}
We use goal-directed
generation of personalized treatments
 optimized for
 expected
 future outcome 
 utility 
 learned from retrospective patient data.
\item \textbf{Testing 
on Diabetes Patient Data:} We demonstrate the potential of our 
framework 
by 
      generating
     insulin treatment 
     and blood glucose 
     trajectories for 
     hospitalized diabetes patients. 
\end{itemize}

\section{Background}
Generating personalized 
treatment trajectories based on historical patient 
data is the ultimate goal of personalized medicine 
with machine learning (ML). 
In this paper, we focus on
the 
generation of multivariate 
insulin strategies 
that lead to optimal expected future blood glucose trajectories (\autoref{fig:overview_generation_trajectories}).
%
Several ML approaches deterministically predict the blood glucose outcome 
\citep{xie2020benchmarking, liu2020predicting,noaro2020machine, jaloli2022long}
and only a few \citep{sergazinov2023gluformer, zhu2023glugan} take into account the inherent uncertainty and multi-modal distribution of the 
future blood glucose. 
Moreover, it is not obvious how these predictions of blood glucose can be utilized for treatment generation.
%
On the other hand, there are several reinforcement-learning approaches
\citep{tejedor2020reinforcement, javad2019reinforcement, shifrin2020near, zhu2020insulin, emerson2023offline}
optimizing 
a deterministic reward function, 
mostly exploiting synthetic or continuously measured blood glucose data, 
and 
considering only a 
simplified action space, for which 
explicit 
constraints have to be introduced to obtain feasible and
consistent treatments.
%
%
We instead try to generate 
feasible multivariate treatment strategies 
consistent over multiple treatments and time without requiring heuristic 
constraints. Moreover,  
we exploit sparse real-world  data from hospitalized diabetes patients 
and optimize a learned probabilistic 
expected utility function. Further, we use deep generative models 
\citep{
Tomczak2022DeepModeling} that learn
the joint distribution over trajectories of treatments and outcomes; 
sharing analogies with 
\citep{ajay2022conditional}.
To the best of our knowledge, 
our approach combining deep generative models with expected utility 
learning is the first to deal with the generation of entire multivariate treatment and outcome trajectories jointly learned 
from retrospective patient data, to achieve feasible, personalized, and optimal treatment strategies.

\begin{figure*}[htp]
\centering
\floatconts
{fig:why_joint}
{\caption{\small
Illustration 
of 
the joint personalized distribution $p(\bs{y}, \bs{x}\vert \bs{c})$, with scalar and non-temporal $\bs{y}$ and $\bs{x}$. 
}}
{\includegraphics[width=0.99\textwidth]{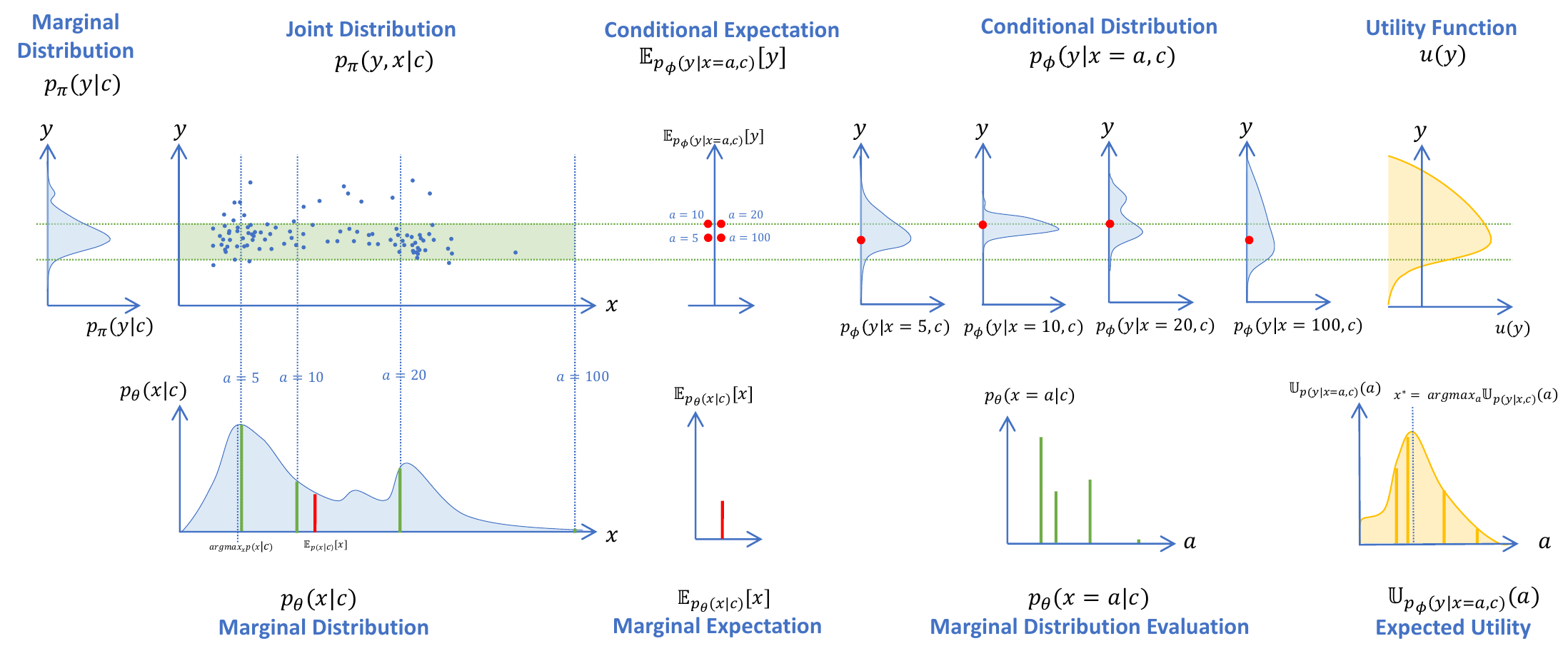}}
\end{figure*}

\section{Methodology}
\label{sec:methodology}

We consider multivariate time series data
involving
outcome
$ \bs{y}_{1:T} \in 
\RR{P \times T}$,
treatment
$ \bs{x}_{1:T} \in 
\mathcal{X} = 
\RR{D \times T}$,
and
covariate
$\bs{v}_{1:T} \in  \RR{V \times T}$ trajectories 
(\autoref{fig:overview_generation_trajectories}). 
%
For insulin treatments, we consider blood glucose as outcome, 
basal and bolus insulin as treatments, and other medications and carbohydrates as covariates.
We divide all trajectories into past and future windows, 
where the latter has
 fixed length $K$, for instance  $ \bs{y}_{1:T} = [\bar{\bs{y}}, \bs{y}]$ with $\bs{y} \in \RR{P \times K}$. 
%
We define the 
 personalized context $\bs{c} = \{\bar{\bs{y}}, \bar{\bs{x}}, \bar{\bs{v}}, \bs{v}, \bs{s} \}$ and 
 with
 non-temporal 
 patient data  $\bs{s}$. 
We use a 
 dataset $\mathcal{D} = \{
 \bs{y}_{1:T_i}^i,  \bs{x}_{1:T_i}^i,  \bs{v}_{1:T_i}^i, \bs{s}^i
 \}_{i=1}^N$ 
 with
 $N$ historical patient 
 time series.



\subsection{Generation of Personalized Treatments}
\label{sec:ddecision}



In this section, we present our approach to leverage 
multivariate time series samples
for 
learning  feasible 
and ``optimal" treatment strategies
based on the 
joint personalized distribution
$p(\bs{y}, \bs{x}  \vert \bs{c}) = p(\bs{y}  \vert \bs{x}, \bs{c}) p(\bs{x}  \vert \bs{c})$, where ``optimality" is defined in the subsequent sections.
In the case of insulin treatments, ``optimal" is achieved when the future outcome (i.e. blood glucose values) lie within a predefined range, where too low values are very dangerous, and too high values are very undesired.

\paragraph{Direct Approach:}
We could directly learn a probabilistic mapping $p(\bs{x}  \vert \bs{c})$  from the past personalized conditions to the future treatments
and 
take 
the most probable treatment strategy
 $$ \bs{x}^*
  =
   \underset{\bs{x}\in \mathcal{X}}{\argmax}~
   ~
   p(\bs{x}  \vert \bs{c}).$$
However, this approach learns to replicate the treatments in the historical training data, 
that might turn problematic if these were sub-optimal 
regarding treatment outcomes.

\paragraph{Indirect Approach:}
In a second approach, we  probabilistically predict the future outcomes $\bs{y} \in \RR{P\times T}$ and 
 specify a 
utility function $u(\bs{y})$ 
describing the
preferences 
for the future treatment outcomes (Section \ref{sec:app:decision}). 
For insulin treatments, the utility function is highest for blood glucose values in the predefined band.
We
can indirectly
optimize the conditional expected utility 
$ \mathbb{U}_{p(\bs{y}  \vert \bs{x}, \bs{c})}(\bs{x})
=
\int
p(\bs{y}\vert \bs{x}, \bs{c})
  u(\bs{y})
d \bs{y}
$ 
(\ref{sec:app:condUT}), that is,
 $$ \bs{x}^*
 =
   \underset{\bs{x} \in \mathcal{X} }{\argmax}
   ~  \mathbb{U}_{p(\bs{y}  \vert \bs{x}, \bs{c})}(\bs{x}),
$$
so that 
optimal
treatments 
regarding
future treatment outcomes are learned.
However, optimizing over all possible treatments $\mathcal{X}=\RR{D \times T}$ 
yields unfeasible treatments, 
e.g.\ 
not adapted to the 
personalized conditions, and the evaluation of the 
learned 
expected utility is not accurate, compare
\autoref{fig:overview_generation_trajectories} (right) and 
\autoref{fig:why_joint}. 

\paragraph{Joint Approach:}
We instead propose
to exploit the joint distribution 
$p(\bs{y}, \bs{x}  \vert \bs{c}) = p(\bs{y}  \vert \bs{x}, \bs{c}) p(\bs{x}  \vert \bs{c})$.
In particular, we first sample $U$ %
feasible
personalized treatment suggestions 
$$\bs{x}^{(1)}_{\vert \bs{c}},\ldots,\bs{x}^{(u)}_{\vert \bs{c}},\ldots, \bs{x}^{(U)}_{\vert \bs{c}}   \overset{\mathrm{iid}}{\sim}
p(\bs{x}\vert  \bs{c})$$
and secondly, we choose the treatment with maximal conditional expected utility 
 $$ \bs{x}^*
  =
   \underset{\bs{x}^{(u)}_{\vert \bs{c}} \sim  p(\bs{x}  \vert \bs{c})}{\argmax}~
    \mathbb{U}_{p(\bs{y}  \vert \bs{x}, \bs{c})}(\bs{x}^{(u)}_{\vert \bs{c}} ).
    $$
%
%
%
%
%
With this approach, we learn feasible and personalized treatment strategies $\bs{x}^{(u)} \sim p(\bs{x}  \vert \bs{c}) $ that have a high probability of resulting in good future treatment outcomes $ \mathbb{U}_{p(\bs{y}  \vert \bs{x}, \bs{c})}(\bs{x}^{(u)})
=
\int
p(\bs{y}\vert \bs{x}^{(u)}, \bs{c})
  u(\bs{y})
d \bs{y}. 
$ 
However, the exact computation of this integral is infeasible for non-trivial distributions
$p(\bs{y}\vert \bs{x}, \bs{c})$, therefore we propose an approach based on deep generative models. 

\paragraph{Expected Utility with Generative Models:}
Deep conditional
generative models are powerful 
for efficiently generating 
conditional samples of complex distributions over multivariate time series, for instance, 
personalized outcome trajectories
$$\bs{y}^{(1)}_{\vert \bs{x}, \bs{c}}, \ldots,\bs{y}^{(s)}_{\vert \bs{x}, \bs{c}}, \ldots, \bs{y}^{(S)}_{\vert \bs{x}, \bs{c}}  \overset{\mathrm{iid}}{\sim}
p(\bs{y}\vert \bs{x}, \bs{c}).$$ These  can be exploited  to approximate the integral 
in the
expected utility
$  \mathbb{U}_{p(\bs{y}  \vert \bs{x}, \bs{c})}(\bs{x})=
   \int
   p(\bs{y}\vert \bs{x}, \bs{c})
  u(\bs{y})
d \bs{y}
$
with  Monte-Carlo samples 
 $$\hat{  \mathbb{U}}_{p(\bs{y}  \vert \bs{x}, \bs{c})}(\bs{x})
  =
\frac{1}{S} \sum_{s=1}^S
  u\left(\bs{y}^{(s)}_{\vert \bs{x}, \bs{c} }\right),$$
leading to the optimization of the sample-based 
expected utility, that is,
    $$
    \bs{x}^*
  =
       \underset{\bs{x}^{(u)} \sim  p(\bs{x}  \vert \bs{c})}{\argmax}~
 \sum_{s=1}^S
  u\left(\bs{y}^{(s)}_{\vert \bs{x}^{(u)}, \bs{c} }\right).$$
  
%
%
%

%
%
%
\subsection{Deep Generative Model}
 \label{sec:gen_model}




In this section, we present our approach to learn the personalized joint distribution 
$$p_{\pi}( \bs{y} , \bs{x} \vert \bs{c} ) = 
 p_{\phi}( \bs{y} \vert \bs{x}, \bs{c} )
    p_{\theta}( \bs{x} \vert \bs{c} )
    $$
  with deep generative time series models \citep{Tomczak2022DeepModeling, murphy2022probabilistic} 
  with
  learnable parameters 
  $\pi = \{\phi, \theta\}$.
In particular,  we focus on a probabilistic encoder-decoder transformer  \citep{vaswani2017attention}, as illustrated in 
\autoref{fig:y_pred} 
and further described in
\ref{sec:app:gen_model}.
%
%

\paragraph{Encoder of Personalized History:}
\label{sec:encoder}
For a deterministic encoder, the personalized history $ \bs{h}$ is mapped to 
 a fixed  latent representation  $\bs{z} = f_{\psi}(  \bs{h} ) \in \RR{L\times (T-K)}$, 
 whereas for a probabilistic encoder,
a deep parametrized probability distribution  
$$
    p_{\psi}( \bs{z} \vert \bs{h} )
    =
    \prod_{t=1}^{T-K}
     \prod_{l=1}^L
    \mathcal{N}\left( 
    z_{tl} \vert 
    \mu_{\psi}^{tl}( \bs{h} ), 
     \sigma_{\psi}^{tl}( \bs{h} )
    \right)
    $$
with deep neural networks 
$ \mu_{\psi}^{tl}( \bs{h} )$
and
$\sigma_{\psi}^{tl}( \bs{h} )$ is learned.

\paragraph{Outcome Trajectory Generation:}
\label{sec:outcome_pred}

The 
multivariate outcomes $\bs{y} \in \RR{P\times K}$
are parametrized as 
$$
    p_{\phi}( \bs{y} \vert \bs{x}, \bs{v}, \bs{z} )
    =
    \prod_{t=1}^K
     \prod_{p=1}^P
    \mathcal{N}\left( 
    y_{tp} \vert 
    \mu_{\phi}^{tp}( \bs{x}, \bs{v}, \bs{z}), 
     \sigma_{\phi}^{tp} 
    \right),
    $$
with learnable
mean $ \mu_{\phi}^{tp}( \bs{x}, \bs{v}, \bs{z})$ and variance
$ \sigma_{\phi}^{tp}
$.

\paragraph{Treatment Trajectory Generation:}
\label{sec:treatment_pred}
For the multivariate treatments
$\bs{x}\in \RR{D \times K}$, 
a Poisson likelihood
$$
    p_{\theta}( \bs{x} \vert \bs{v}, \bs{z} )
    =
      \prod_{t=1}^K
     \prod_{d=1}^D
     \mathcal{P}\left( 
    x_{td} \vert 
    \lambda_{\theta}^{tp}( \bs{x}, \bs{v}, \bs{z})
     \right)
     $$
with deep parametrized mean 
$ \lambda_{\theta}^{tp}( \bs{x}, \bs{v}, \bs{z})$ is used. 

\subsubsection{Source of Stochasticity}
\label{sec:objective_fun}

Although the likelihoods above are parametric distributions, the predictive distributions of the outcome 
and treatment 
trajectories can be arbitrarily complex when using deep generative models.
We compare our model with three different sources of stochasticity in the generative process as illustrated in \autoref{fig:prob_mode}.
%
%
\begin{figure}[htbp]
\centering
\floatconts
{fig:prob_mode}
{\caption{Probabilistic Modes 
of Generative Model.}}
{\includegraphics[width=0.9\linewidth]{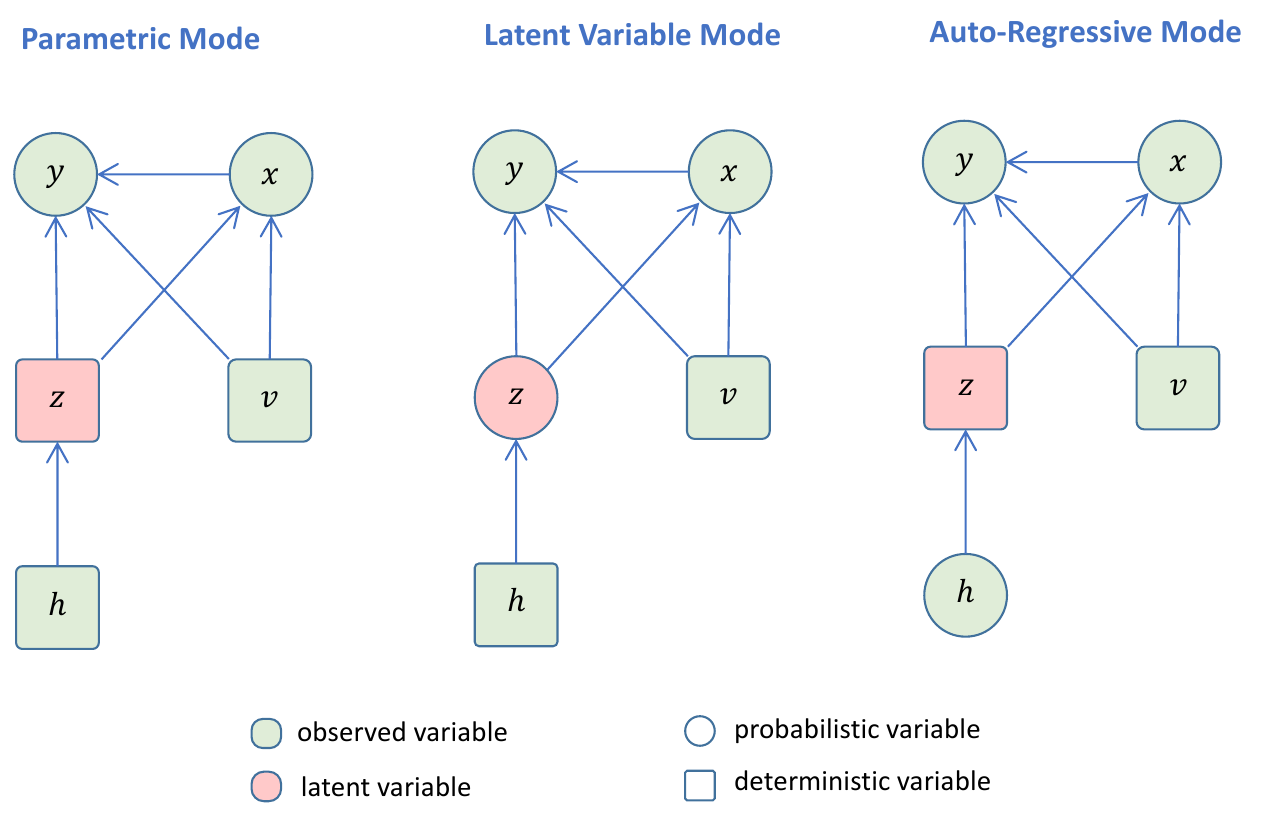}}
\end{figure}
\paragraph{Parametric Mode:}
For a deterministic encoder 
without any additional stochasticity,
the joint distribution 
$
p_{\pi}( \bs{y} , \bs{x}\vert \bs{v}, \bs{z})$
is
$
    p_{\phi}( \bs{y} \vert \bs{x}, \bs{v}, \bs{z})
    p_{\theta}( \bs{x} \vert \bs{v}, \bs{z})
    $
with encoded history
$\bs{z} = f_{\psi}(\bs{h})$.
Training with
the log-likelihood 
$$\mathcal{L}_1(\pi; \mathcal{D}) =
\log
p_{\pi}( \bs{y} , \bs{x} \vert \bs{v}, \bs{z} )$$
can be used as a baseline 
with resulting 
parametric predictive distributions (factorized Gaussians and Poissons) of 
treatments and outcomes.

\begin{figure*}[htp!]
\centering
\floatconts
{fig:y_pred}
{\caption{\small
Transformer-based model (left), illustration (middle), and evaluation (right) of 
personalized prediction. 
}}
{\includegraphics[width=0.8\linewidth]{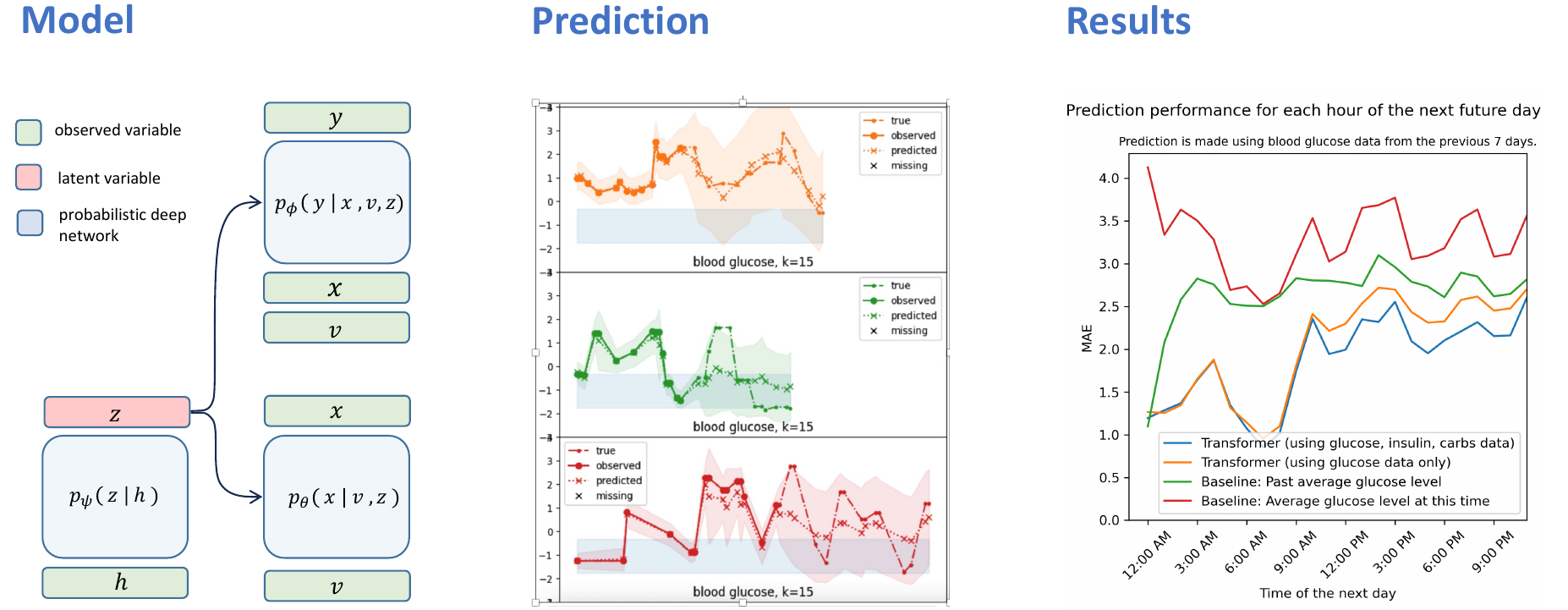}}
\end{figure*}

\paragraph{Latent Variable Mode:}
 With a probabilistic encoder 
 $p_{\psi}( \bs{z} \vert \bs{h} )$
 and corresponding 
 prior 
$ p( \bs{z} ) = \mathcal{N}(\bs{z}\vert \bs{0}, \sigma^2_p \mathbb{I})$ of the latent variable,
the joint distribution is 
$p_{\pi}( \bs{y} , \bs{x}, \bs{z} \vert \bs{v}, \bs{h} )
=
    p_{\phi}( \bs{y} \vert \bs{x}, \bs{v}, \bs{z} )
    p_{\theta}( \bs{x} \vert \bs{v}, \bs{z} )
     p( \bs{z} ).$
Based on 
variational 
inference (\ref{sec:app:laten_var}),
the
objective function is
\begin{align*}
\mathcal{L}_2(\pi; \mathcal{D}) 
&
=
\mathbb{E}_{ p_{\psi}( \bs{z} \vert \bs{h} )}\left[
\log  p_{\phi}( \bs{y} \vert \bs{x}, \bs{v}, \bs{z} )
\right]
\\
&
+
\mathbb{E}_{ p_{\psi}( \bs{z} \vert \bs{h} )}\left[
\log   p_{\theta}( \bs{x} \vert \bs{v}, \bs{z} )
\right]
\\
&
-
KL
\left[
p_{\psi}( \bs{z} \vert \bs{h} )
\vert \vert
 p( \bs{z} )
\right],
\end{align*}
leading to infinite mixtures of Gaussian and Poisson
predictive distributions of the outcomes and treatments.

\paragraph{Auto-Regressive Mode:}
%
In the auto-regressive case,
the joint distribution
$p_{\pi}( \bs{y} , \bs{x}\vert \bs{v}, \bs{z} )$
is decomposed as
$
\prod_{t=1}^T
p_{\pi}( \bs{y}_t , \bs{x}_t \vert \bs{v}_t, \bs{z}_{1:t-1} ) 
    $
with 
varying encoded
history
$\bs{z}_{1:t-1} = f_{\psi}([\bs{y}_{1:t-1} , \bs{x}_{1:t-1}, \bs{v}_{1:t-1}, \bs{s} ])$.
Maximizing the log-likelihood yields
\begin{align*}
\mathcal{L}_3(\pi; \mathcal{D})
=
\sum_{t=1}^T 
&
\log p\left( \bs{y}_t \vert \bs{x}_t, \bs{v}_t, \bs{z}_{1:t-1} \right)
\\
+ &\log
    p\left( \bs{x}_t \vert \bs{v}_t, \bs{z}_{1:t-1} \right)
\end{align*}
allows us to generate multivariate trajectories beyond infinite mixture distributions.


\paragraph{Implementation:}
We use a transformer encoder \citep{vaswani2017attention} with multi-head self-attention blocks to encode the personalized history $p_{\psi}( \bs{z} \vert \bs{h} )$.
The output  $p_{\phi}( \bs{y} \vert \bs{x}, \bs{v}, \bs{z} )$ and the treatment $p_{\theta}( \bs{x} \vert \bs{v}, \bs{z} )$ prediction networks are implemented using two transformer decoders with self-attention and cross-attention blocks to attend also to the encoded past $\bs{z}$ (\autoref{fig:y_pred}).
%
We train our model 
with
historical patient trajectories 
$\mathcal{D} = \{\mathcal{D}_i\}_{i=1}^N$
using the three 
objective functions
$\mathcal{L}_1$, 
$\mathcal{L}_2$, 
and
$\mathcal{L}_3$ 
, see also
\ref{sec:implementation-details}, \ref{sec:app:optimization}.

\section{Experiments and Results}
\label{sec:results}

\paragraph{Insulin and Blood Glucose Trajectories:}
%
We test our approach using retrospective patient data from University hospital ($N = 2621$) comprised of blood glucose trajectories $ \bs{y}_{1:T} \in  \RR{1 \times T}$,
multivariate
treatment strategies $ \bs{x}_{1:T} \in  \RR{2 \times T}$ representing 
 basal and bolus insulin injection doses,
 and carbohydrates as covariates $\bs{v}_{1:T} \in  \RR{1 \times T}$.
 We plan to extend the treatment strategies and covariates to other medications in the future. For our heterogeneous  patient data, 
 we provide comprehensive descriptive data analysis 
in Appendix \ref{sec:app:descriptive}. Note that this dataset presents particular challenges due to the infrequent reporting of carbohydrates and sparse blood glucose measurements, typically only 3-4 times per day.

\paragraph{Results for Blood Glucose Prediction:}
We provide preliminary results 
from our proposed method
for generating personalized insulin treatment and future outcome blood glucose trajectories.
In particular, 
the probabilistic prediction of the personalized outcomes trajectories $ p( \bs{y} \vert \bs{x}, \bs{c} )$ are shown in  \autoref{fig:y_pred} (middle and right), and in the Appendix 
in \autoref{tab:y_performance_model_comparison_table} and \autoref{fig:online}.
%
%
From \autoref{fig:y_pred} (right), we observe that our transformer model 
outperforms two simple baselines for predicting blood glucose 24 hours ahead.
The green line refers to the baseline using the patient's past average glucose level. The red baseline is the average of glucose measurements at a particular time point
from all patients.
For the sake of simplicity,  the prediction always starts at midnight,
so that 
the best performance gap
is 
around 3-6 hours in the future. In future work, we will predict the outcomes 
from any 
time point,
and improve the experiments in several directions, as outlined in
Appendix \ref{sec:app:pl_ex}.

\section{Conclusion}
\label{sec:conclusion}

In this 
paper, we propose a framework for leveraging conditional samples of multivariate time series generated by  deep conditional generative 
time series 
models.
%
It can be used to learn the complex interaction between outcome, treatment, and covariate trajectories from retrospective patient time series data. 
Moreover, feasible, personalized, and optimal treatment trajectories can be generated by combining it with conditional expected utility maximization.
Our preliminary results focused on modeling insulin and blood glucose trajectories from diabetes patients, however, 
our framework can be generally applied to other healthcare patient trajectory data.
We plan to improve our work 
with several 
experiments, as outlined in
Appendix \ref{sec:app:pl_ex}.
Moreover, 
we pursue
interesting extensions for our framework following the end-to-end goal-directed generation of personalized treatment strategies, as outlined in \ref{sec:app:improved}.
%



\bibliography{jmlr-sample}

\appendix

\section{Details and Extension about Model}\label{sec:app:details}

\subsection{Notation}
\label{sec:app:notation}

In this section, we provide more precise notation.
In particular, we consider multivariate time series data, including outcome
$ \bs{y}_{1:T_y} \in \mathcal{Y} = \RR{P \times T_y}$,
treatment
$ \bs{x}_{1:T_y} \in \mathcal{X} = \RR{D \times T_x}$,
and
covariate
$\bs{v}_{1:T_v} \in  \RR{V \times T_v}$ trajectories.
For each time series, we assume irregularly sampled times
$\bs{\tau}^y_{1:T_y}$, $\bs{\tau}^x_{1:T_x}$, and $\bs{\tau}^v_{1:T_v}$, respectively.
For the sake of simplicity, we assume a fixed future time window of length $K \in \mathbb{N}$ (e.g.\ 24 hours) and divide the trajectories into past and future, 
that is,
$ \bs{y}_{1:T_y} = [\bar{\bs{y}}, \bs{y}]$, $ \bs{x}_{1:T_x} = [\bar{\bs{x}}, \bs{x}]$, and
$ \bs{v}_{1:T_v} = [\bar{\bs{v}}, \bs{v}]$, where
$\bs{y} \in \RR{P \times K_y}$, $\bs{x} \in \RR{D \times K_x}$, and $\bs{v} \in \RR{V \times K_v}$,
compare also \autoref{fig:overview_generation_trajectories} on the left.
Note that, in the implementation, we consider a moving future window.
We define the 
 personalized context $\bs{c} = \{\bar{\bs{y}}, \bar{\bs{x}}, \bar{\bs{v}}, \bs{v}, \bs{s} \}$, where
 $\bs{s}$ are some additional non-temporal patient information.
 We assume a dataset $\mathcal{D} = \{\mathcal{D}_i\}_{i=1}^N$ consisting of $N$ historical patient time series
 $\mathcal{D}_i = \{\bs{y}_{1:T_y^i}^i,  \bs{x}_{1:T_x^i}^i,  \bs{v}_{1:T_v^i}^i, \bs{s}^i\}$, however, we omit the explicit dependency to $i$ if it is clear from the context.

\subsection{Decision Making}
\label{sec:app:decision}

\subsubsection{Expected Utility}

Suppose we have an utility function $u(\bs{y}): \mathcal{Y} \rightarrow \RR{}$  for a random variable $\bs{y} \in 
\mathcal{Y}
$ with density
 $p(\bs{y})$. We define the expected utility $\mathbb{U}_{p(\bs{y})} \in \RR{}$ as
\begin{align*}
  \mathbb{U}_{p(\bs{y})}
  =
  \mathbb{E}_{p(\bs{y})}\left[
  u(\bs{y})
  \right]
  =
   \int
   p(\bs{y})
  u(\bs{y})
d \bs{y}.
\end{align*}

\subsubsection{Conditional Expected Utility}
\label{sec:app:condUT}
Similarly, for a conditional distribution $p(\bs{y}\vert \bs{x})$ conditioned on
$\bs{x} \in 
\mathcal{X}
$,
we define the  conditional expected  utility $ \mathbb{U}_{p(\bs{y}  \vert \bs{x})}(\bs{x})$ as
\begin{align}
\label{eq:ECU}
  \mathbb{U}_{p(\bs{y}  \vert \bs{x})}(\bs{x})
  =
  \mathbb{E}_{p(\bs{y}\vert \bs{x})}\left[
  u(\bs{y})
  \right]
  =
   \int
   p(\bs{y}\vert \bs{x})
  u(\bs{y})
d \bs{y},
\end{align}
which can be considered as a function in $\bs{x}$.

\subsubsection{Joint Expected Utility}
For a joint distribution $p(\bs{y}, \bs{x})$,
we define the  joint expected  utility $ \mathbb{U}_{p(\bs{y}, \bs{x})}$ as
\begin{align*}
  \mathbb{U}_{p(\bs{y} , \bs{x})}
  &=
  \mathbb{E}_{p(\bs{y}, \bs{x})}\left[
  u(\bs{y}, \bs{x})
  \right]
  \\
  &
  =
   \int
   p(\bs{y}, \bs{x})
  u(\bs{y}, \bs{x})
d \bs{y}
~d \bs{x}
\\
&=
  \int
   p(\bs{x}) p(\bs{y} \vert \bs{x})
  u(\bs{y}, \bs{x})
d \bs{y}
~d \bs{x}.
\end{align*}
Here, we introduce the joint utility function 
$u(\bs{y}, \bs{x}): \mathcal{Y}\times \mathcal{X} \rightarrow \RR{}$, which also expresses the utility for the conditional variable $\bs{x}$ (treatment strategies). In the case of a factorized joint utility function
$u(\bs{y}, \bs{x}) = u(\bs{y}) u(\bs{x})$, we can write
\begin{align*}
  \mathbb{U}_{p(\bs{y} , \bs{x})}
&=
  \int
   p(\bs{x}) p(\bs{y} \vert \bs{x})
  u(\bs{y})
   u(\bs{x})
d \bs{y}
~d \bs{x}
\\
&=
   \mathbb{E}_{p( \bs{x})}\left[
     u(\bs{x})
     \mathbb{U}_{p(\bs{y}  \vert \bs{x})}(\bs{x})
    \right],
\end{align*}
which can be decomposed as the expectation over the conditional variables (treatments) of the weighted conditional expected utility.
For a utility function $u(\bs{y}, \bs{x}) = u(\bs{y})$, this simplifies to 
 $\mathbb{U}_{p(\bs{y} , \bs{x})}
=
   \mathbb{E}_{p( \bs{x})}\left[
     \mathbb{U}_{p(\bs{y}  \vert \bs{x})}(\bs{x})
    \right]$.

\subsubsection{Hierarchical Conditional Utility}
We can also extend the introduced concepts to more variables, for instance, conditioning on a context variable 
$\bs{c}$, so that the expected utilities become
$\mathbb{U}_{p(\bs{y}\vert \bs{c})}(\bs{c})$,
$  \mathbb{U}_{p(\bs{y}  \vert \bs{x},  \bs{c})}(\bs{x}, \bs{c})$, and
 $\mathbb{U}_{p(\bs{y} , \bs{x} \vert \bs{c})}(\bs{c})$, which are all functions in $\bs{c}$.

\subsubsection{Multivariate Utility Functions}
 Suppose $\mathcal{Y} =\RR{P}$. We can define a factorized utility function $u(\bs{y}): \RR{P} \rightarrow \RR{}$ 
 \begin{align*}
 u(\bs{y})
  =
 \prod_{p=1}^P
 \alpha_p
 u(y_p)
\end{align*}
with weights $\alpha_p$ for each dimension. This is particularly interesting when modeling 
concurring outcomes, for instance, the success of treatment together with complications or side effects.
This allows specifying the different utilities for multiple outcomes
 and enables the analysis of treatments that satisfy Pareto-optimality regarding multiple outcomes.

\subsubsection{Temporal Expected Utility}
Suppose $\mathcal{Y} =\RR{P \times T}$, that is a temporal indexed multivariate random variable $\bs{y}=\bs{y}_{1:T}\in \RR{P \times T}$. We can directly define 
the utility function over the whole matrix. For instance,
\begin{align*}
 u(\bs{y}_{1:T})
  =
 \sum_{t=1}^T
  \gamma^t u(\bs{y}_t) 
    =
 \sum_{t=1}^T
   \gamma^t 
 \prod_{p=1}^P
\alpha_p u(y_{pt}) 
\end{align*}
with time decaying parameter
$0 < \gamma \leq 1$.
In this case, the expected utility 
becomes
\begin{align}
\label{eq:temp_utility}
\begin{split}
  \mathbb{U}_{p(\bs{y}_{1:T})}
  &=
  \mathbb{E}_{p(\bs{y}_{1:T})}\left[
  u(\bs{y}_{1:T})
  \right]
  \\
  &=
   \int
   p(\bs{y}_{1:T})
  u(\bs{y}_{1:T})
d \bs{y}_{1:T}
  \\
  &=
   \int
   p(\bs{y}_{1:T})
  \sum_{t=1}^T
  \gamma^t u(\bs{y}_t) 
d \bs{y}_{1:T}.
\end{split}
\end{align}
If we assume temporal 
independence, that is, $p(\bs{y}_{1:T}) = \prod_{t=1}^T p(\bs{y}_{t})$, this leads to
$$ \mathbb{U}_{p(\bs{y}_{1:T})}
=
\int
\sum_{t=1}^T
   p(\bs{y}_{t})
  \gamma^t u(\bs{y}_t) 
d \bs{y}_{t}
=
\sum_{t=1}^T
\gamma^t 
 \mathbb{U}_{p(\bs{y}_{t})}.
$$


\subsubsection{Approximation by Generative Models}
\label{sec:app:aprox_gen}
All these integrals are often hard to compute for interesting objects such as complex multivariate time series. On the other hand, deep generative probabilistic models are powerful for generating conditional samples from very complex distributions over objects including multivariate time series efficiently. Thus, we propose to  
 generate multiple conditional samples 
$$\bs{y}^{(1)}_{\vert \bs{x}},\ldots,\bs{y}^{(s)}_{\vert \bs{x}}, \ldots, \bs{y}^{(S)}_{\vert \bs{x}}\sim 
p(\bs{y}\vert \bs{x})$$ 
with deep generative models to obtain a Monte-Carlo-based approximation for the expected utilities. For instance, the
expected utility 
  $  \mathbb{U}_{p(\bs{y}  \vert \bs{x})}(\bs{x})=
   \int
   p(\bs{y}\vert \bs{x})
  u(\bs{y})
d \bs{y}
$ in \autoref{eq:ECU} is approximated by
\begin{align*}
 \hat{  \mathbb{U}}_{p(\bs{y}  \vert \bs{x})}(\bs{x})
  =
\frac{1}{S} \sum_{s=1}^S
  u(\bs{y}^{(s)}_{\vert \bs{x}}).
\end{align*}
In the case of multivariate time series $\bs{y}=\bs{y}_{1:T}\in \RR{P \times T}$,
we jointly sample  entire trajectories 
$[\bs{y}_{1}^{(s)}, \ldots, \bs{y}_{T}^{(s)}] = \bs{y}_{1:T}^{(s)}\sim 
p(\bs{y}_{1:T})$,
so that
 the sampling version of \autoref{eq:temp_utility} becomes 
\begin{align*}
 \hat{  \mathbb{U}}_{p(\bs{y}_{1:T})  }
  =
  \frac{1}{S} \sum_{s=1}^S
  u(\bs{y}^{(s)}_{1:T})
  =
\frac{1}{S} \sum_{s=1}^S
 \sum_{t=1}^T
  \gamma^t 
  u(\bs{y}^{(s)}_t).
\end{align*}
Similarly, we can estimate 
$\hat{  \mathbb{U}}_{p(\bs{y}_{1:T} \vert \bs{x}_{1:T})  }$
for conditioning on a whole treatment trajectory
$\bs{x}=\bs{x}_{1:T}\in \RR{D \times T}$.

Moreover, we want to emphasize that
 we can take gradients with respect to the parameters of the deep parametrized distributions based on end-to-end optimization. Sampling can be included via the re-parametrization trick.

 \subsubsection{Exact Computation}
\label{sec:app:exact_comp}
Under certain circumstances, the expected utility 
$
  \mathbb{U}_{p(\bs{y}  \vert \bs{x})}(\bs{x})
  =
  \mathbb{E}_{p(\bs{y}\vert \bs{x})}\left[
  u(\bs{y})
  \right]
  =
   \int
   p(\bs{y}\vert \bs{x})
  u(\bs{y})
d \bs{y}
$
in Equation \eqref{eq:ECU} can be computed exactly. It can be used as baseline to compare the approximation with deep generative models as discussed in the previous section.
For instance, if we assume a joint Gaussian distribution or a Gaussian Process (GP) \citep{williams2006gaussian, schurch2020recursive, schurch2023correlated} for $p(\bs{y}\vert \bs{x})$ and 
a simple utility function $u(\bs{y})$ (for instance affine, exponential, periodic, Gaussian, log-Gaussian, ...), then the expected utility can be computed exactly. This has the benefit that the variance from the two-stage sampling of the response and treatment can be reduced by only sampling the treatments.

\subsubsection{Improved Treatment Generation }
\label{sec:app:improved}

In order to directly learn a better distribution 
$ p_{\theta}(\bs{x}  \vert \bs{c})$ which take into account desired properties regarding optimal treatment responses,
we propose to learn the 
\begin{align*}
  \theta^*
 & =
   \underset{  \theta }{\argmax}~  \mathbb{U}_{p(\bs{y}, \bs{x} \vert \bs{c})}
   \\
   & =
   \underset{  \theta }{\argmax}~  
   \int
   p_{\theta}(\bs{x} \vert \bs{c}) p(\bs{y} \vert \bs{x}, \bs{c})
  u(\bs{y})
d \bs{y}
~d \bs{x},
\end{align*}
so that we can then directly take the
most probable treatment 
\begin{align*}
  \bs{x}^*
  =
   \underset{\bs{x}}{\argmax}~ p_{\theta^*}(\bs{x}  \vert \bs{c})
\end{align*}
or a few high-probability samples  from
$ p_{\theta^*}(\bs{x}  \vert \bs{c}),$
which are now coupled so that they satisfy also the utility constraints.
This idea is based on the work of \cite{mollaysa2020goal}.

This approach can also be combined with log-likelihood-based inference, so that 
the objective function becomes
\begin{align*}
  \theta^*
 & =
   \underset{  \theta }{\argmax}~  \alpha \mathbb{U}_{p(\bs{y}, \bs{x} \vert \bs{c})}
   + (1-\alpha)
   \mathcal{L}(\theta; \mathcal{D})
   \\
   & =
   \underset{  \theta }{\argmax}~  
  \alpha \int
   p_{\theta}(\bs{x} \vert \bs{c}) p(\bs{y} \vert \bs{x}, \bs{c})
  u(\bs{y})
d \bs{y}
~d \bs{x}
\\
&+
(1-\alpha)
\log
p_{\theta}( \bs{y} , \bs{x} \vert \bs{v}, f(\bs{h}) ),
\end{align*}
where we showed the parametric mode for the sake of simplicity.

\subsubsection{Learning Utility for x}
\label{sec:app:learn_ut_x}

Alternatively, we could also learn the parameters $\pi$ of the utility function $ u_{\pi}(\bs{x})$ over the treatments
$\bs{x}$ with deep learning, that is,
\begin{align*}
  \pi^*
 & =
   \underset{  \pi }{\argmax}~  \mathbb{U}_{p(\bs{y}, \bs{x} \vert \bs{c})}
   \\
   & =
   \underset{  \pi }{\argmax}~  
   \int
   p_{\theta}(\bs{x} \vert \bs{c}) p(\bs{y} \vert \bs{x}, \bs{c})
  u(\bs{y})
    u_{\pi}(\bs{x})
d \bs{y}
~d \bs{x},
\end{align*}
which constitutes a pragmatic approach for generating efficient treatment suggestions that result in favorable outcomes.

\subsection{Deep Generative Model}
\label{sec:app:gen_model}

In this section, we present more details of our approach to learn the conditional joint distribution 
$$p_{\pi}( \bs{y} , \bs{x} \vert \bs{c} ) = 
 p_{\phi}( \bs{y} \vert \bs{x}, \bs{c} )
    p_{\theta}( \bs{x} \vert \bs{c} )
    $$
  with deep generative time series models \citep{Tomczak2022DeepModeling, murphy2022probabilistic} from retrospective patient trajectories data, where we introduce
  the learnable parameters 
  $\pi = \{\phi, \theta\}$.
    Among the many choices for deep generative models,
    a common property is to   
    implicitly learning the joint distribution with deep neural networks by providing a mechanism to generate samples from these distributions.
In particular,  we focus on an encoder-decoder architecture based on a transformer \citep{vaswani2017attention} as illustrated in 
\autoref{fig:y_pred} on the left.

\subsubsection{Encoder of Personalized History}
\label{sec:app:encoder}
We consider either a deterministic or probabilistic encoder $p_{\psi}( \bs{z} \vert \bs{h} )$. In the former, the personalized history $ \bs{h}$ is mapped to 
 a fixed  latent representation  $\bs{z} = f_{\psi}(  \bs{h} ) \in \RR{L\times (T-K))}$, whereas in the latter
we learn a deep parametrized probability distribution  
\begin{align*}
    p_{\psi}( \bs{z} \vert \bs{h} )
    =
    \prod_{t=1}^{T-K}
     \prod_{l=1}^L
    \mathcal{N}\left( 
    z_{tl} \vert 
    \mu_{\psi}^{tl}( \bs{h} ), 
     \sigma_{\psi}^{tl}( \bs{h} )
    \right),
\end{align*}
with deep neural networks 
$ \mu_{\psi}^{tl}( \bs{h} )$
and
$\sigma_{\psi}^{tl}( \bs{h} )$.

\subsubsection{Outcome Predictor}
\label{sec:app:outcome_pred}

The future multivariate outcome trajectory $\bs{y} \in \RR{P\times K}$
is parametrized as 
\begin{align*}
    p_{\phi}( \bs{y} \vert \bs{x}, \bs{v}, \bs{z} )
    =
    \prod_{t=1}^K
     \prod_{p=1}^P
    \mathcal{N}\left( 
    y_{tp} \vert 
    \mu_{\phi}^{tp}( \bs{x}, \bs{v}, \bs{z}), 
     \sigma_{\phi}^{tp} 
    \right),
\end{align*}
where the mean $ \mu_{\phi}^{tp}( \bs{x}, \bs{v}, \bs{z})$ and variance
$ \sigma_{\phi}^{tp}
$
are learned with neural networks. Although this likelihood is a parametric distribution, the final distribution of the outcomes can be arbitrarily complex, 
see Section
\ref{sec:objective_fun}.

\subsubsection{Treatment Predictor}
\label{sec:app:treatment_pred}
For the treatment strategies
$\bs{x}\in \RR{D \times K}$, we assume a Poisson likelihood
\begin{align*}
    p_{\theta}( \bs{x} \vert \bs{v}, \bs{z} )
    =
      \prod_{t=1}^K
     \prod_{d=1}^D
     \mathcal{P}\left( 
    x_{td} \vert 
    \lambda_{\theta}^{tp}( \bs{x}, \bs{v}, \bs{z})
     \right),
\end{align*}
with deep parametrized mean function 
$ \lambda_{\theta}^{tp}( \bs{x}, \bs{v}, \bs{z})$. 
Note that the final distribution of the deep generative model is non-parametric, 
see next section.

\subsubsection{Source of Stochasticity}
\label{sec:app_objective_fun}

We compare our model in three different modes as illustrated in \autoref{fig:prob_mode} for modeling the source of the 
stochasticity in the generation of outcome  $\bs{y}$ and treatment $\bs{x}$ trajectories. As a baseline method, we  use a deterministic encoder and a simple parametric distribution for the outcomes $\bs{y}$ and treatments $\bs{x}$ without any additional stochasticity in the model, so that the the predictive distributions are actually Gaussian and Poisson distributed as defined
in Sections 
\ref{sec:app:outcome_pred} and \ref{sec:app:treatment_pred}. However, these distributions are too restrictive and
do not match real-world data (Appendix \ref{sec:app:descriptive}). As a second mode, we use a latent variable model with a probabilistic encoder (\ref{sec:app:encoder}), so that when sampling from the latent representation, we get an infinite mixture of Gaussian or a mixture of Poisson distributions, allowing to generate rather complex distributions.
The third mode is based on auto-regressive learning and sampling, where we use $K=1$ and plug back the previous values to the history $\bs{h}$, which is then the source of stochasticity in the sampling. This allows to sample multivariate trajectories beyond infinite mixture distributions.

\paragraph{Parametric Mode:}
For a deterministic encoder, we have the joint distribution
\begin{align*}
p_{\pi}( \bs{y} , \bs{x}\vert \bs{v}, f_{\psi}(\bs{h}) )
=
    p_{\phi}( \bs{y} \vert \bs{x}, \bs{v},f_{\psi}(\bs{h}) )
    p_{\theta}( \bs{x} \vert \bs{v}, f_{\psi}(\bs{h}))
\end{align*}
and the objective function 
is the log-likelihood
\begin{align*}
\mathcal{L}_1(\pi; \mathcal{D}) =
\log
p_{\pi}( \bs{y} , \bs{x} \vert \bs{v}, f_{\psi}(\bs{h}) ).
\end{align*}

\paragraph{Latent Variable Mode:}
\label{sec:app:laten_var}

In the latent variable mode, we have the joint distribution
\begin{align*}
p_{\pi}( \bs{y} , \bs{x}, \bs{z} \vert \bs{v}, \bs{h} )
=
    p_{\phi}( \bs{y} \vert \bs{x}, \bs{v}, \bs{z} )
    p_{\theta}( \bs{x} \vert \bs{v}, \bs{z} )
     p( \bs{z} )
\end{align*}
with prior distribution
$ p( \bs{z} ) = \mathcal{N}(\bs{z}\vert \bs{0}, \sigma^2_p \mathbb{I})$.
The ideal objective function would be the  \textit{marginal} log-likelihood
$\mathcal{M}(\pi; \mathcal{D}) =
\log
\int
p_{\pi}( \bs{y} , \bs{x}, \bs{z} \vert \bs{v}, \bs{h} )
d \bs{z},$
which is not feasible to compute exactly. Therefore, we propose to use an amortized variational inference approach
\citep{kingma2013auto, blei2017variational}, which maximizes a lower bound
$\mathcal{L}_2(\pi; \mathcal{D}) \leq \mathcal{M; \mathcal{D}}(\pi)$,
that is
\begin{align*}
\mathcal{L}_2(\pi; \mathcal{D}) 
&=
\mathbb{E}_{ p_{\psi}( \bs{z} \vert \bs{h} )}\left[
\log  p_{\phi}( \bs{y} \vert \bs{x}, \bs{v}, \bs{z} )
\right]
\\
&+
\mathbb{E}_{ p_{\psi}( \bs{z} \vert \bs{h} )}\left[
\log   p_{\theta}( \bs{x} \vert \bs{v}, \bs{z} )
\right]
\\
&-
KL
\left[
p_{\psi}( \bs{z} \vert \bs{h} )
\vert \vert
 p( \bs{z} )
\right].
\end{align*}

\paragraph{Auto-Regressive Mode:}

In the auto-regressive mode, we set the window size to $K=1$ and we decompose the joint distribution
as
\begin{align*}
&p_{\pi}( \bs{y} , \bs{x}\vert \bs{v}, f_{\psi}(\bs{h}) )
\\
=
&\prod_{t=1}^T
p_{\pi}( \bs{y}_t , \bs{x}_t \vert \bs{v}_t, f_{\psi}(\bs{h}_{1:t-1})  )
\\
=
&\prod_{t=1}^T
    p_{\phi}\left( \bs{y}_t \vert \bs{x}_t, \bs{v}_t,f_{\psi}(\bs{h}_{1:t-1}) \right)
    p_{\theta}\left( \bs{x}_t \vert \bs{v}_t, f_{\psi}(\bs{h}_{1:t-1}) \right),
\end{align*}
where
we define the temporal varying personalized history
$\bs{h}_{1:t-1} = [\bs{y}_{1:t-1} , \bs{x}_{1:t-1}, \bs{v}_{1:t-1}, \bs{s} ]$.
By maximizing the log-likelihood, we get the objective
\begin{align*}
\mathcal{L}_3(\pi; \mathcal{D})& =
\log 
\prod_{t=1}^T
p_{\pi}( \bs{y}_t , \bs{x}_t \vert \bs{v}_t, f_{\psi}(\bs{h}_{1:t-1})  )
\\
&
=
\sum_{t=1}^T 
\log p_{\phi}\left( \bs{y}_t \vert \bs{x}_t, \bs{v}_t,f_{\psi}(\bs{h}_{1:t-1}) \right)
\\
&+ \log
    p_{\theta}\left( \bs{x}_t \vert \bs{v}_t, f_{\psi}(\bs{h}_{1:t-1}) \right).
\end{align*}
In the training, we can use the actually observed values $\bs{h}_{1:t-1} = [\bs{y}_{1:t-1} , \bs{x}_{1:t-1}, \bs{v}_{1:t-1}, \bs{s} ]$ without any stochasticity, which is called \textit{teacher mode}. Only for the generation of novel trajectories, we auto-regressively sample from the model.

\paragraph{Optimization:}
\label{sec:app:optimization}
In all three modes $\mathcal{L}_1(\pi; \mathcal{D})$, $\mathcal{L}_2(\pi; \mathcal{D})$, $\mathcal{L}_3(\pi; \mathcal{D})$, we optimize the
parameters  with a datasets
$\mathcal{D} = \{\mathcal{D}_i\}_{i=1}^N$ 
by
$$\pi^* = 
\argmax_{\pi}
\sum_{i=1}^N
\sum_{k=0}^{T_i-K}
\mathcal{L}^i( \pi; \mathcal{D}_i^k ),
$$
which is computed with stochastic optimization using mini-batches of patients. 
Here, $\mathcal{D}_i^k$ refers to the data of patient $i$ until time point $k$.

\subsection{Implementation Details}\label{sec:implementation-details}
\subsubsection{Data Pre-processing}
Since the time series is irregularly sampled, we convert it to an hourly sampled time series via imputation. The glucose level $\bs{x}$ is linearly imputed. The treatment $\bs{x}$ (i.e., basal and bolus insulin dose) is imputed with zero.
We linearly scale $\bs{y}$ to have mean $0$ and standard deviation $1$. We also linearly scale $\bs{x}$ to have standard deviation $1$.
\subsubsection{Transformer architecture}
We use the transformer encoder, decoder, and positional encoding from \cite{vaswani2017attention}. We train with learning rate $10^{-3}$, batch size $8$, one decoder and encoder layer, $64$ features, $16$ heads, and a single-hidden-layer embedding of size $100$ for the feedforward network of the encoder/decoder as well as for the embedding of the input variables. When training the outcome predictor $p_{\phi}( \bs{y} \vert \bs{x}, \bs{v}, \bs{z} )$, we only compute the loss for the non-imputed measured future $\bs{y}$ values.

\subsection{Planned Experiments}
\label{sec:app:pl_ex}
In this Section,
we provide several planned experiments for our method.

In general,
in order to assess the quality of our approach, we will compare the prediction accuracy and reliability (uncertainty) of the outcome and treatments against the true future in the retrospective data. Moreover, to assess the quality of the generated samples, we will implement a simple classifier to distinguish whether it is a real or generated sample.

\subsubsection{Outcome Trajectory Generation}

In our model, we will examine how the personalized history $\bs{h}$, the covariates $\bs{v}$ and the treatment $\bs{x}$ affect the prediction performance for $\bs{y}$. In particular, we will compare different conditional sets such as
$p_{\phi}( \bs{y} )$,
$p_{\phi}( \bs{y} \vert  \bs{\tau}_{\bs{y} } )$,
$p_{\phi}( \bs{y} \vert  \bar{ \bs{y} } )$,
$p_{\phi}( \bs{y} \vert  \bar{ \bs{x} } )$,
$p_{\phi}( \bs{y} \vert  \bs{h} )$,
$p_{\phi}( \bs{y} \vert \bs{x}, \bs{h} )$,
$p_{\phi}( \bs{y} \vert  \bs{v}, \bs{h} )$,
and 
$p_{\phi}( \bs{y} \vert \bs{x}, \bs{v}, \bs{h} )$.

\subsubsection{Treatments Strategy Generation}
Similarly, we compare the samples of different distributions of treatment trajectories, that is
$p_{\theta}( \bs{x} )$, 
$p_{\theta}( \bs{x} \vert \bs{v})$,
$p_{\theta}( \bs{x} \vert  \bs{h})$,
and 
$p_{\theta}( \bs{x} \vert \bs{v}, \bs{h})$.
It is particularly interesting to see the effect of the past history $\bs{h}$
as well as the effect from the carbohydrates $\bs{v}$.

\subsubsection{Comparison of Modes}
We will compare the three different modes of stochasticity in the generative model, as explained in Section \ref{sec:objective_fun}. 
It will be interesting to see whether the parametric, the latent variable, or the auto-regressive approach leads to the best performance.
Besides comparing the performance, it will be particularly interesting to compare the quality of the generated multivariate samples of the trajectories. 
Moreover, we will 
examine the latent space of the latent variable approach. In particular, we try to visualize the learned latent space of the treatment trajectories and check if the interpolation property is satisfied. Moreover, we check if we can generate samples around particular interesting points in the latent space.


\subsubsection{Comparison with other Approaches}
We will compare our approach with other prediction models for the outcomes and treatments, such as deterministic deep neural networks (RNN, CNN).
Moreover, it would be interesting to find a setting in which we can compare our generated treatment with approaches from reinforcement learning and compare their quality. In particular, it would be interesting to compare how realistic and consistent our multivariate samples are.

\subsubsection{Generalization to other Applications}
We plan to test our general framework for generating personalized and optimal treatment strategies to other diseases. For instance, we plan to apply it to cancer treatment data as well as rheumatic disease data
\citep{trottet2023generative}.

\subsection{Descriptive Data Analysis}
\label{sec:app:descriptive}

In the following, we describe the blood glucose, carbohydrate intake, basal insulin, and bolus insulin data. We plot their typical values/doses (throughout the day) as well as the number of observations/insulin injections per day.
We would like to empathize that the shown results show global distributions involving all patients, whereas the personalized distributions of the trajectories of treatment outcomes are mostly multi-modal, often due to the latent, unobserved variables such as carbohydrates or medications.

\begin{figure*}[htbp]
    \centering
    \floatconts
    {fig:blood_glucose_simple_histograms}
    {\caption{Blood glucose is mostly measured around 3-4 times per day and usually takes values between 5 and 15.}}
    {\includegraphics[width=0.8\linewidth]{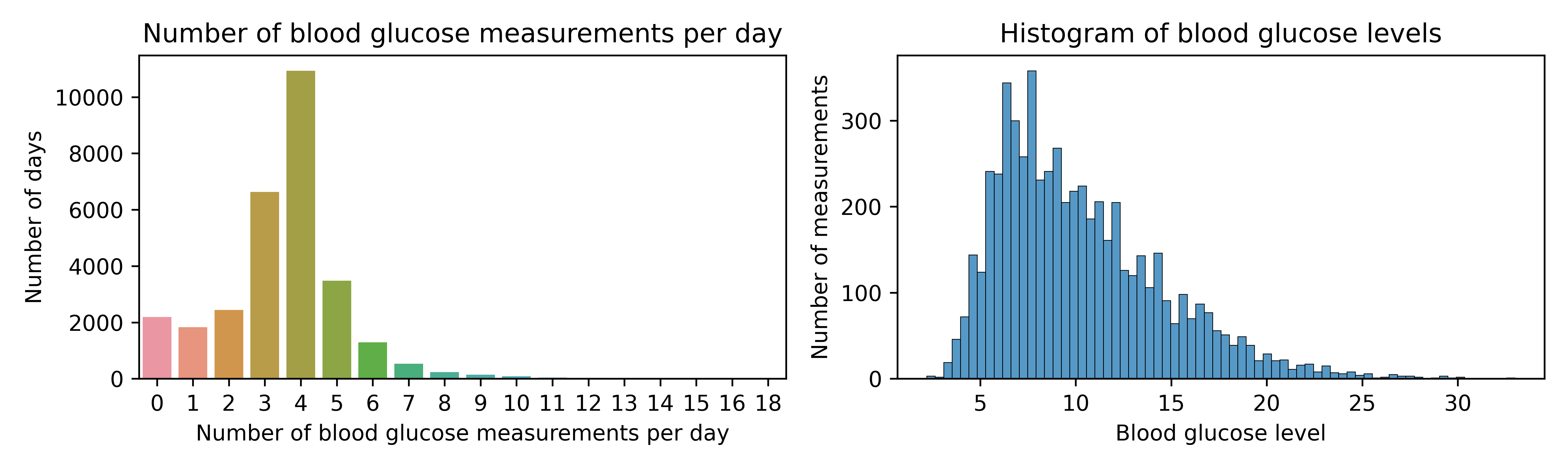}}
\end{figure*}
\begin{figure*}[htbp]
    \centering
    \floatconts
    {fig:blood_glucose_number_measurements_by_time_of_day}
    {\caption{Blood glucose is usually measured at 7-8 a.m., 12  p.m., 6  p.m., or 10  p.m..}}
    {\includegraphics[width=0.7\linewidth]{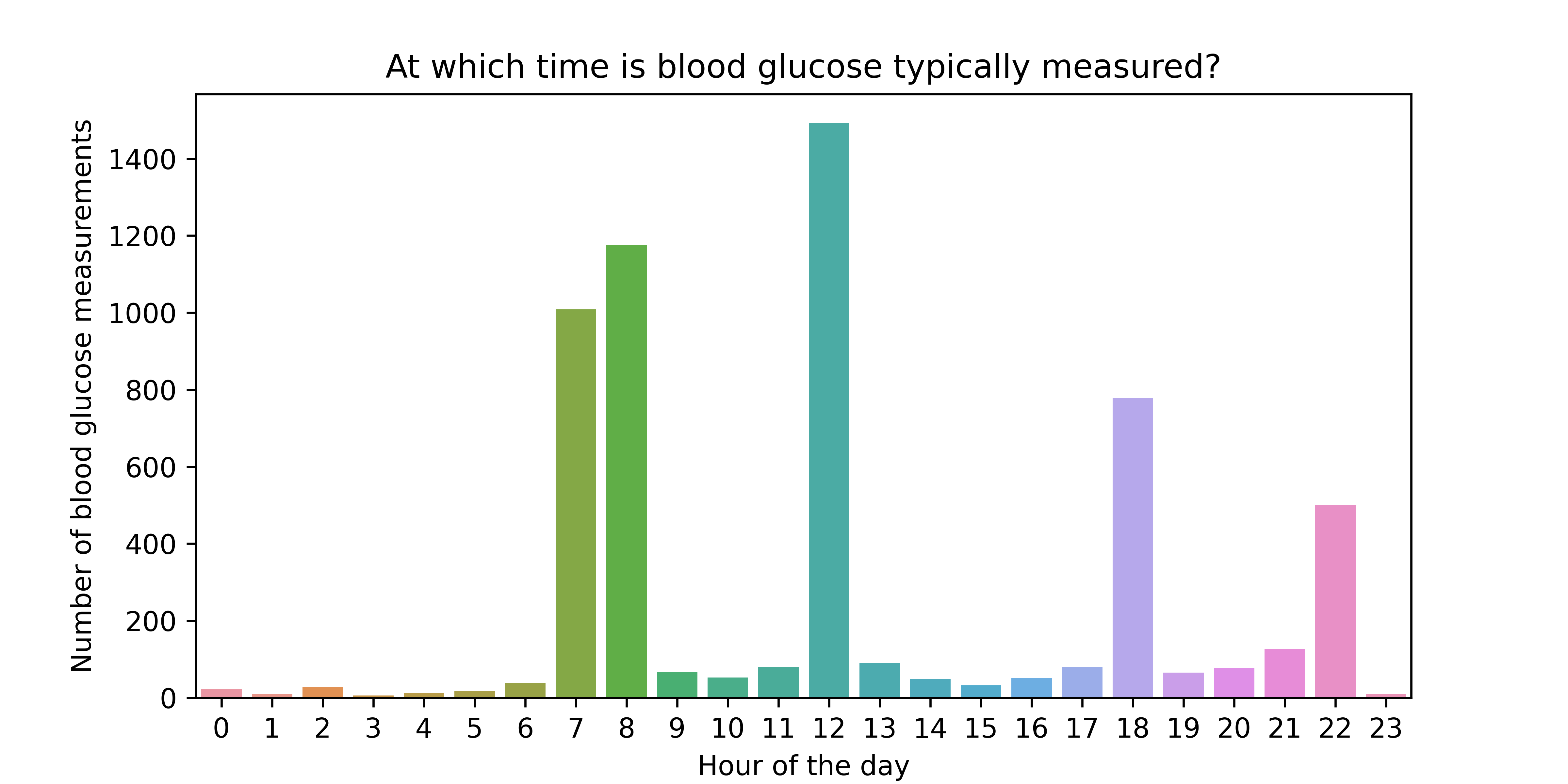}}
\end{figure*}
\begin{figure*}[htbp]
    \centering
    \floatconts
    {fig:blood_glucose_quantile_values}
    {\caption{Blood glucose levels exhibit fluctuations throughout the day, with their lowest point typically occurring around 7 a.m.}}
    {\includegraphics[width=0.7\linewidth]
    {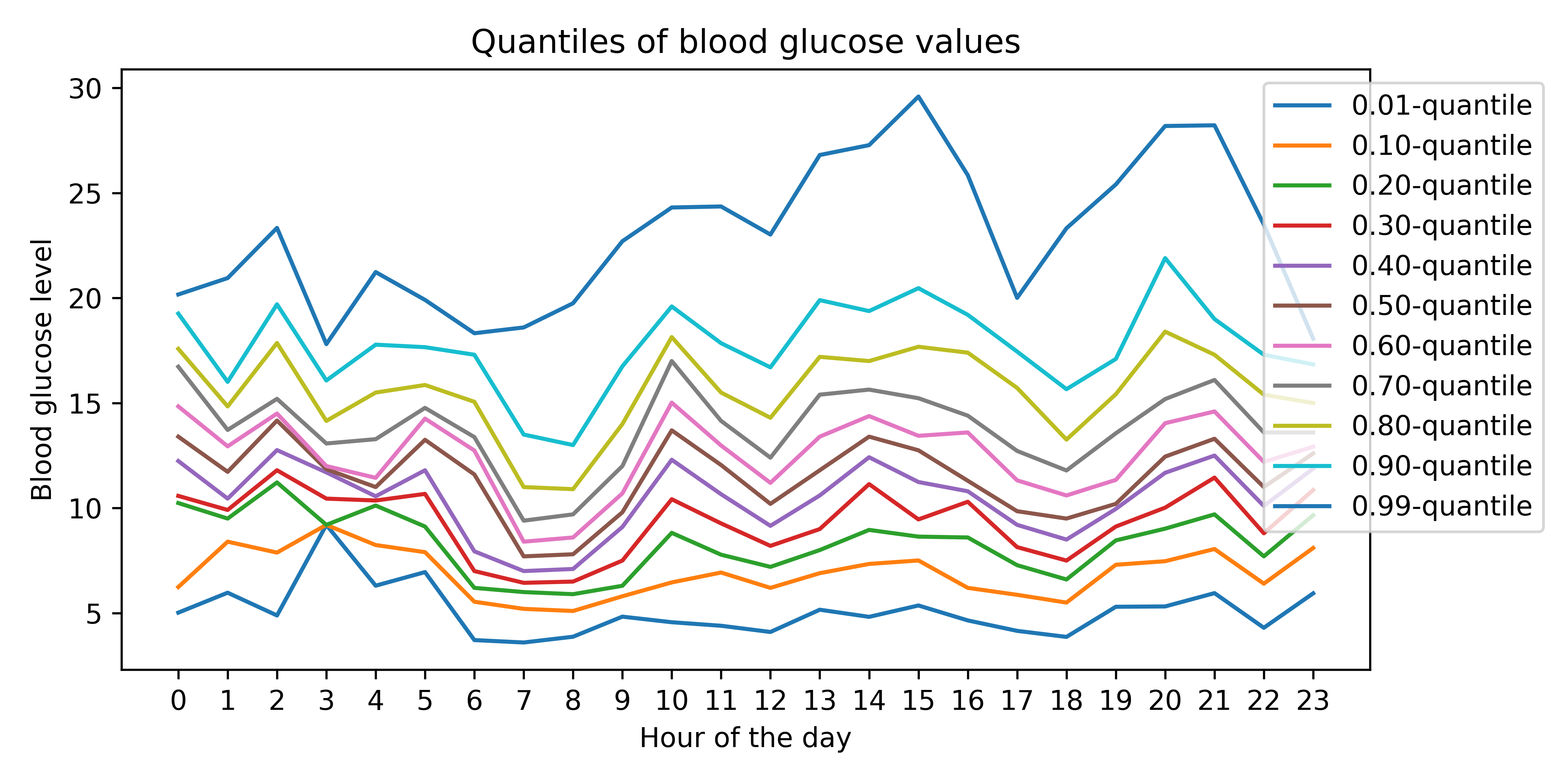}}
\end{figure*}

\begin{figure*}[ht]
    \centering
    \floatconts
    {fig:carbs_simple_histograms}
    {\caption{Carbohydrate intake data is usually not provided. If it is provided, then usually 3 times per day. Carbohydrate consumption per meal mostly ranges from 20 to 70. }}
    {\includegraphics[width=0.8\linewidth]{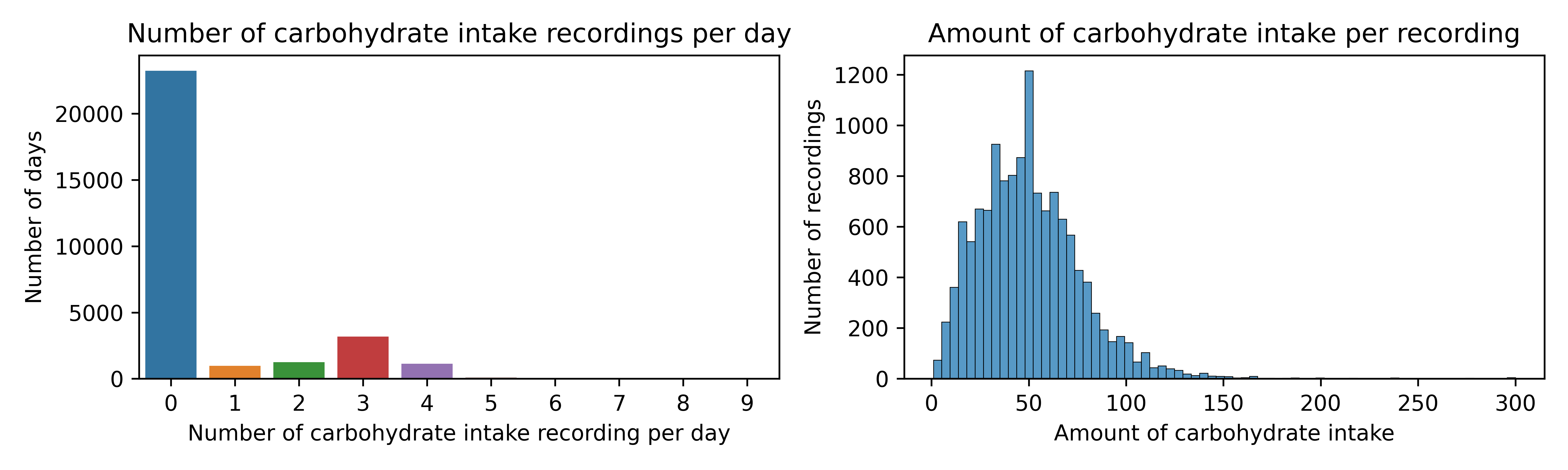}}
\end{figure*}
\begin{figure*}[ht]
    \centering
    \floatconts
    {fig:carbs_number_measurements_by_time_of_day}
    {\caption{Carbohyrdates are usually consumed at 7-8 a.m., 12  p.m., or 6  p.m..}}
    {\includegraphics[width=0.7\linewidth]{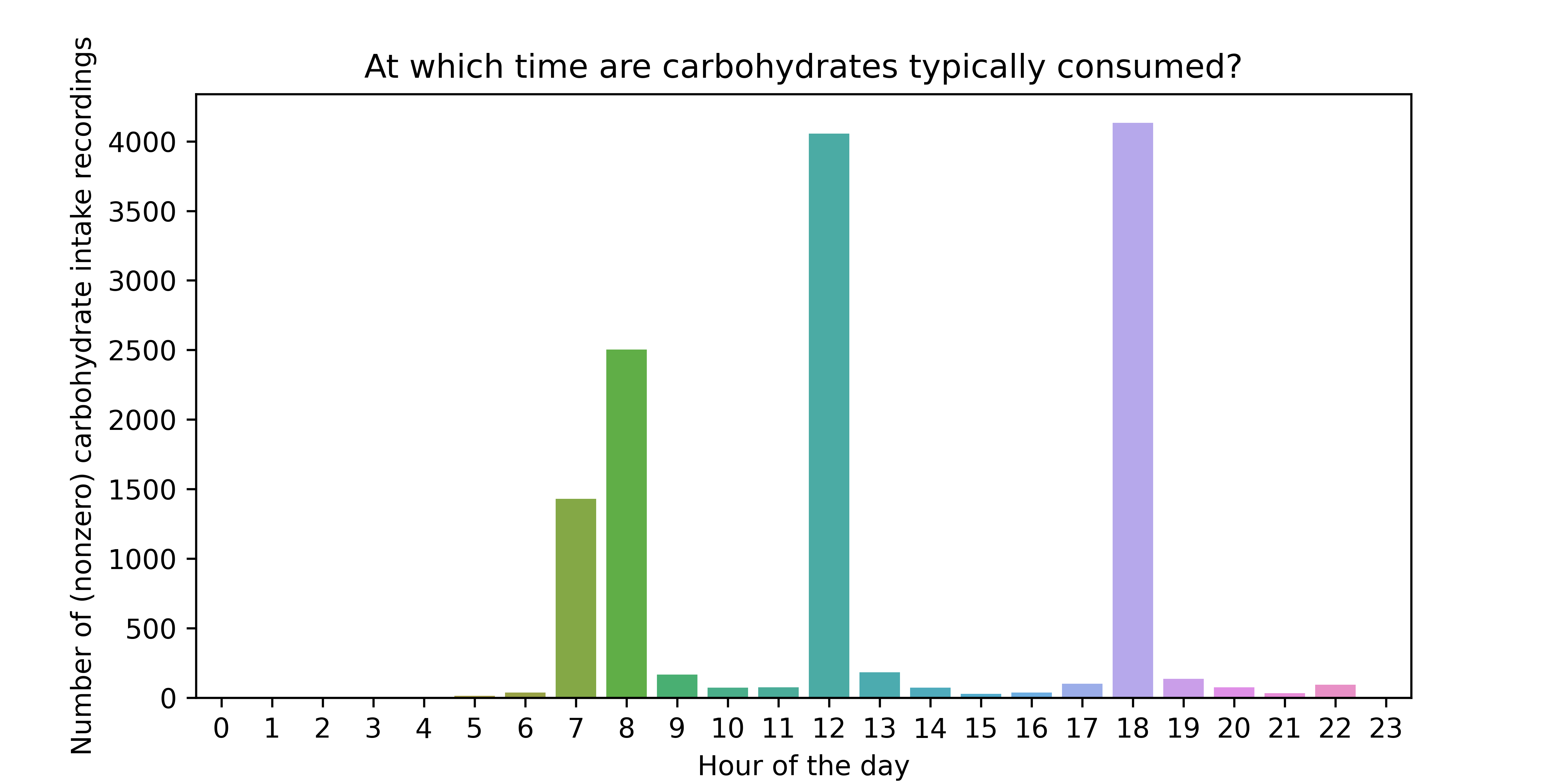}}
\end{figure*}
\begin{figure*}[ht]
    \centering
    \floatconts
    {fig:carbs_quantile_values}
    {\caption{Carbohydrate intake throughout the day. If carbohydrates are consumed and this is reported, then the amount of carbohydrates does not depend on the time of the day.}}
    {\includegraphics[width=0.7\linewidth]{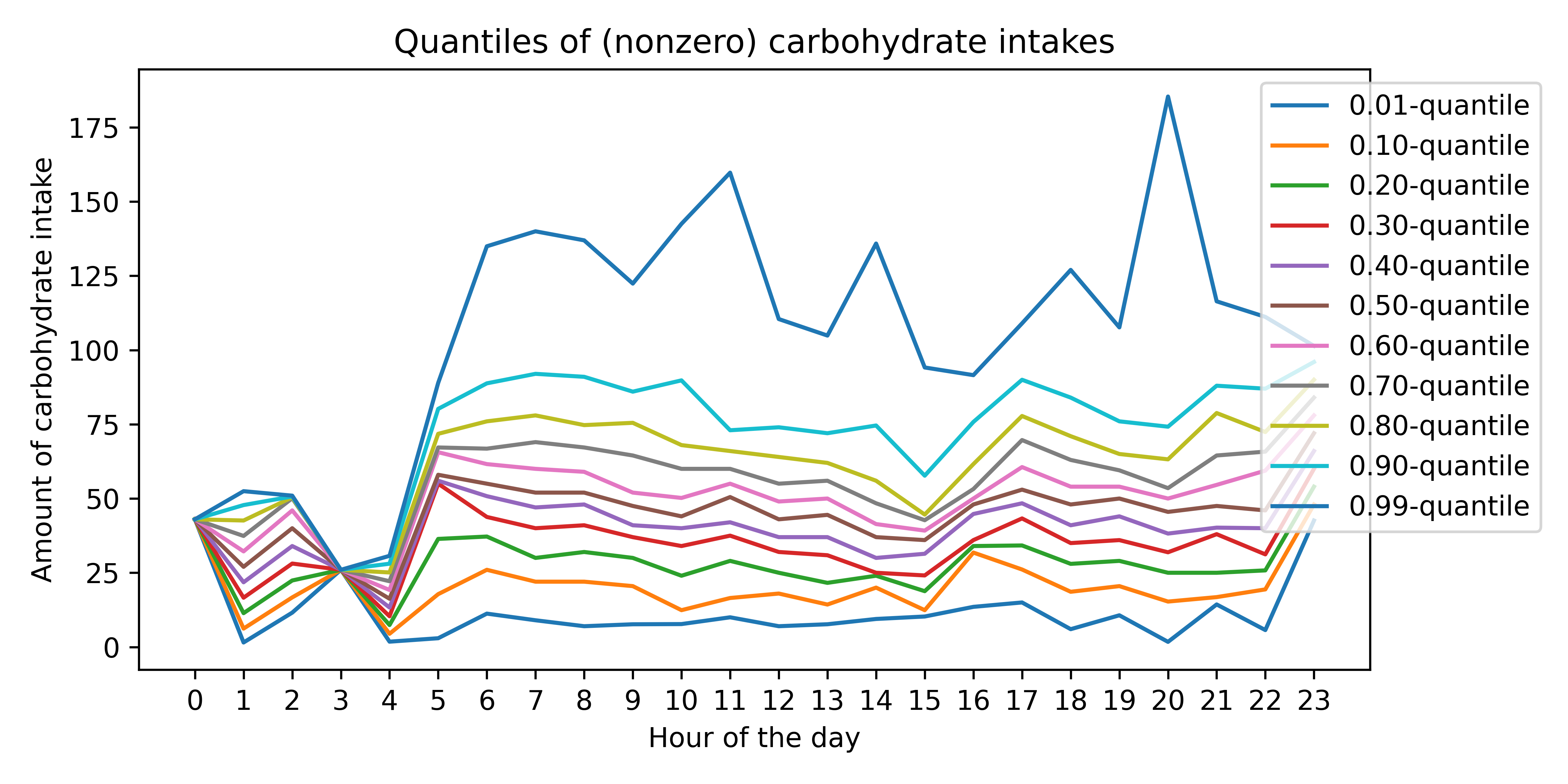}}
\end{figure*}

\begin{figure*}[htbp]
    \centering
    \floatconts
    {fig:basal_insulin_simple_histograms}
    {\caption{Basal insulin is usually injected once per day or not at all. The dose per injection is around 2-20.}}
    {\includegraphics[width=0.8\linewidth]{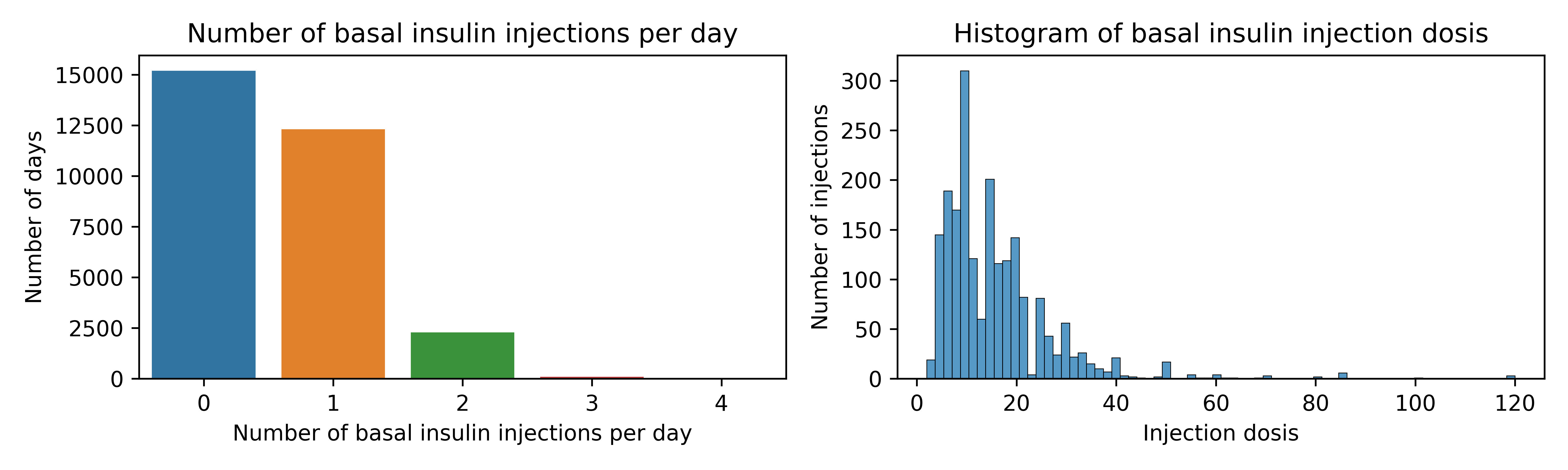}}
\end{figure*}
\begin{figure*}[htbp]
    \centering
    \floatconts
    {fig:basal_insulin_number_injections_by_time_of_day}
    {\caption{Basal insulin injection almost exclusively occurs at 7–8 a.m. or 6 p.m.}}
    {\includegraphics[width=0.7\linewidth]{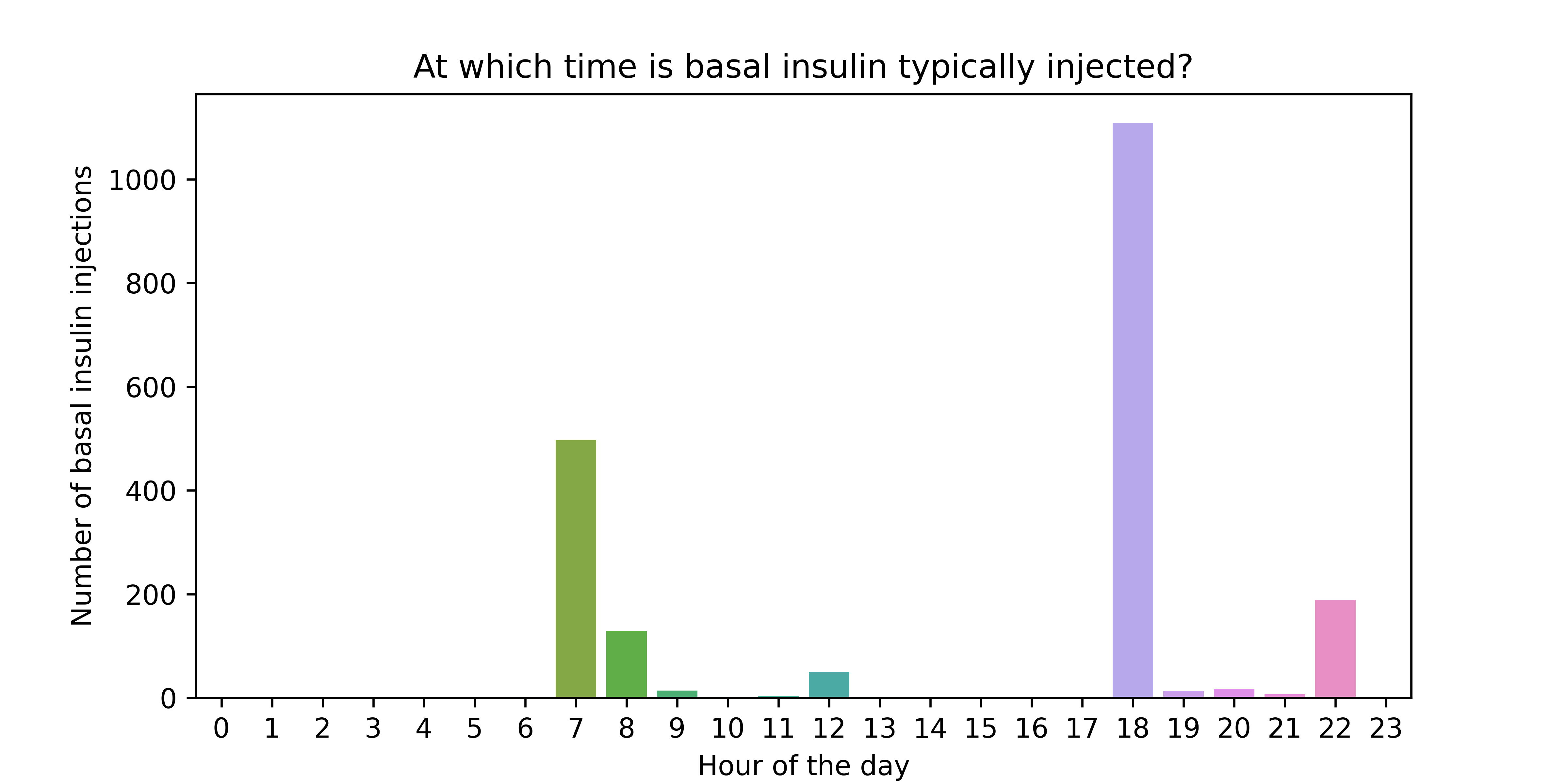}}
\end{figure*}
\begin{figure*}[htbp]
    \centering
    \floatconts
    {fig:basal_insulin_quantile_injection_dosis}
    {\caption{The basal insulin injection dose is independent of the time of the day.}}
    {\includegraphics[width=0.7\linewidth]{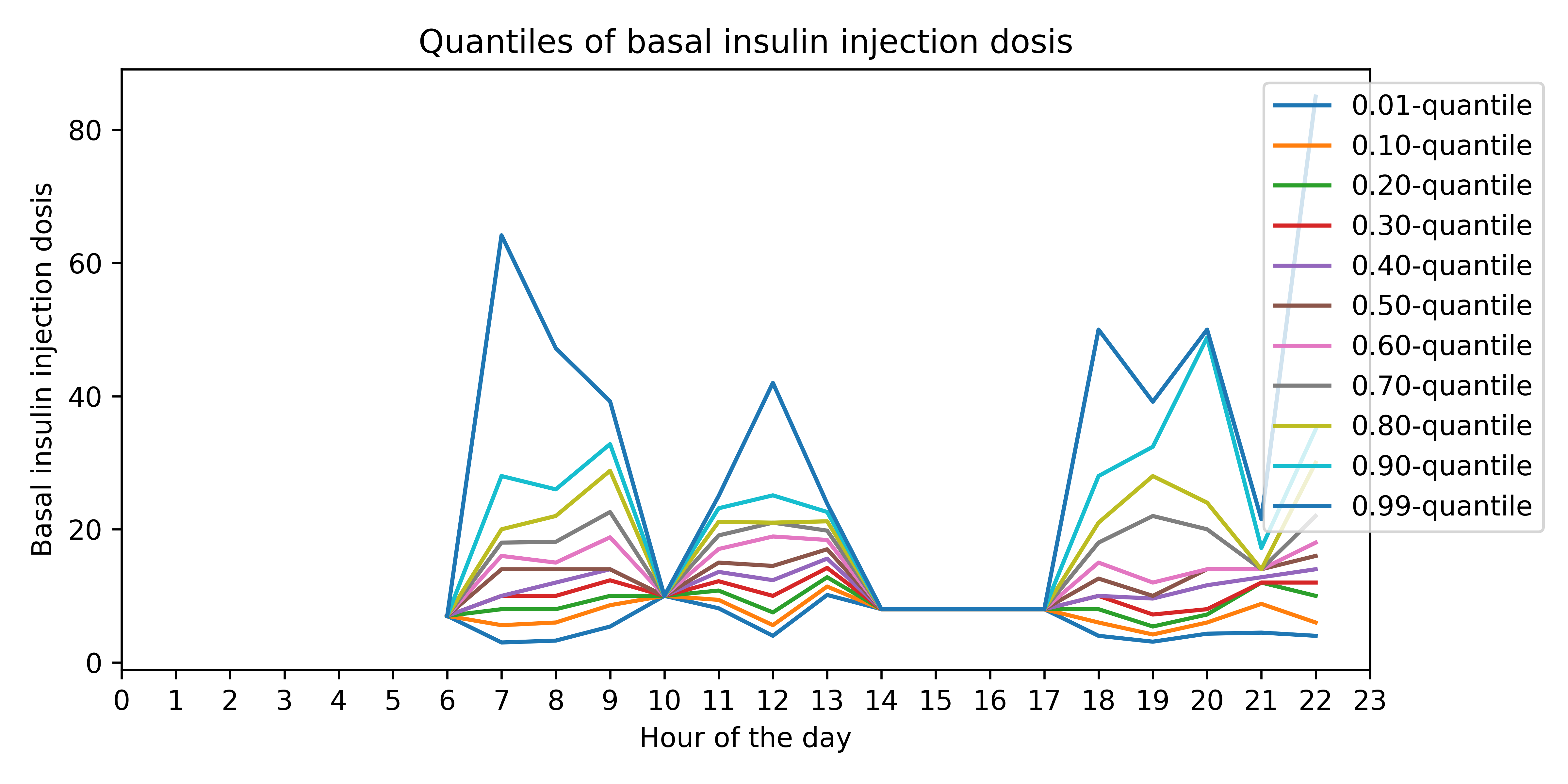}}
\end{figure*}

\begin{figure*}[htbp]
    \centering
    \floatconts
    {fig:bolus_insulin_simple_histograms}
    {\caption{Bolus insulin is usually injected 0–4 times per day. The dose per injection is around 0–30.}}
    {\includegraphics[width=0.8\linewidth]{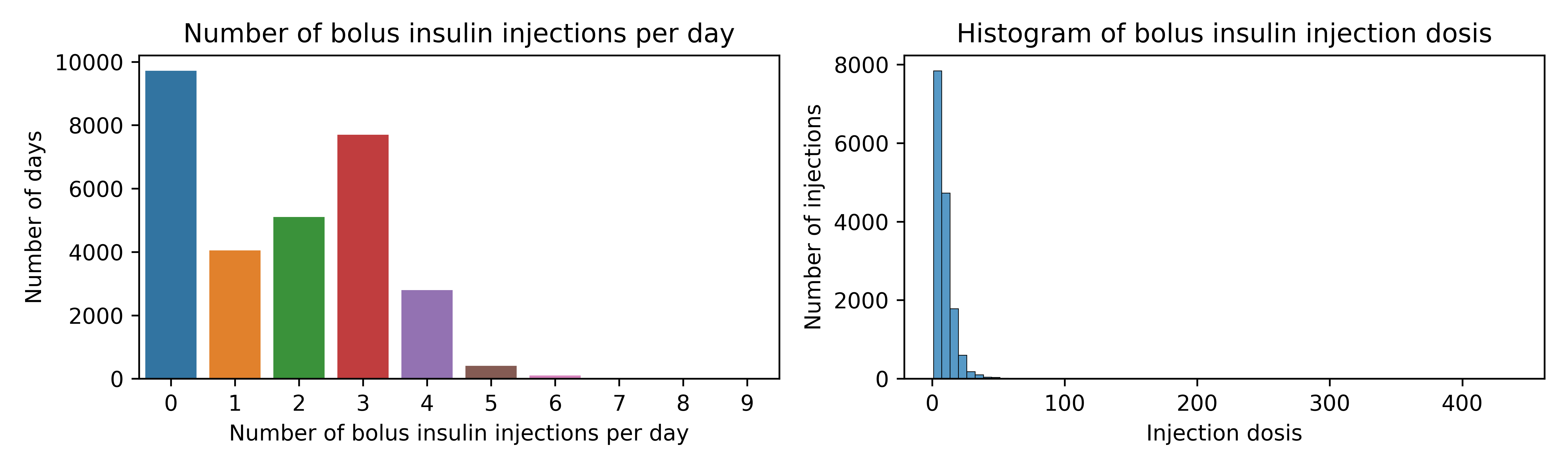}}
\end{figure*}
\begin{figure*}[htbp]
    \centering
    \floatconts
    {fig:bolus_insulin_number_injections_by_time_of_day}
    {\caption{If bolus insulin is injected, then this occurs at 7–8 a.m., 12 a.m., 6 p.m., and, less frequently, at 10  p.m.}}
    {\includegraphics[width=0.7\linewidth]{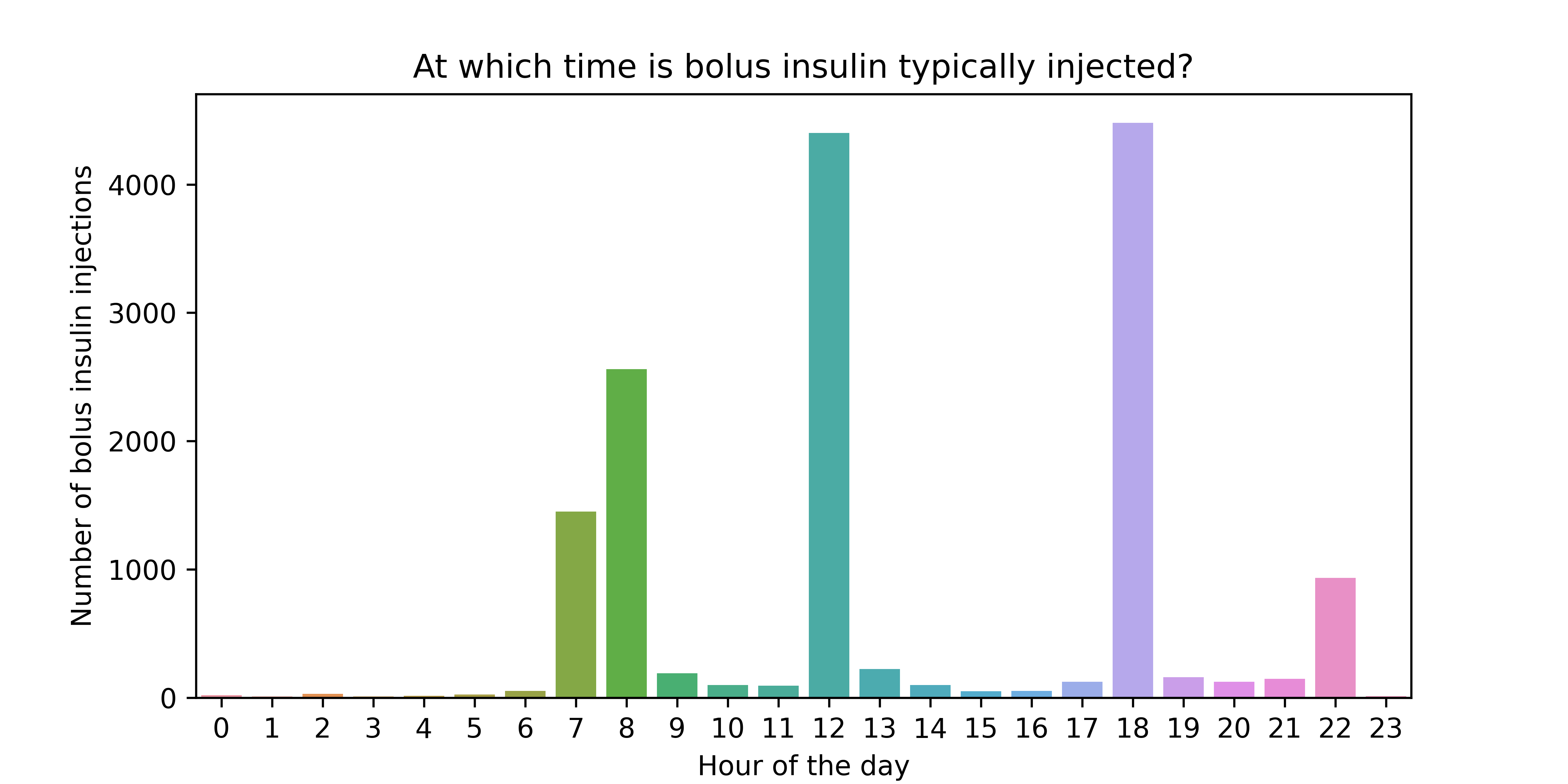}}
\end{figure*}
\begin{figure*}[htbp]
    \centering
    \floatconts
    {fig:bolus_insulin_quantile_injection_dosis}
    {\caption{Bolus insulin injection doses throughout the day. Injections at 10 p.m. almost always have the same low dose. The injection dose at other times is higher and varies more.}}
    {\includegraphics[width=0.7\linewidth]{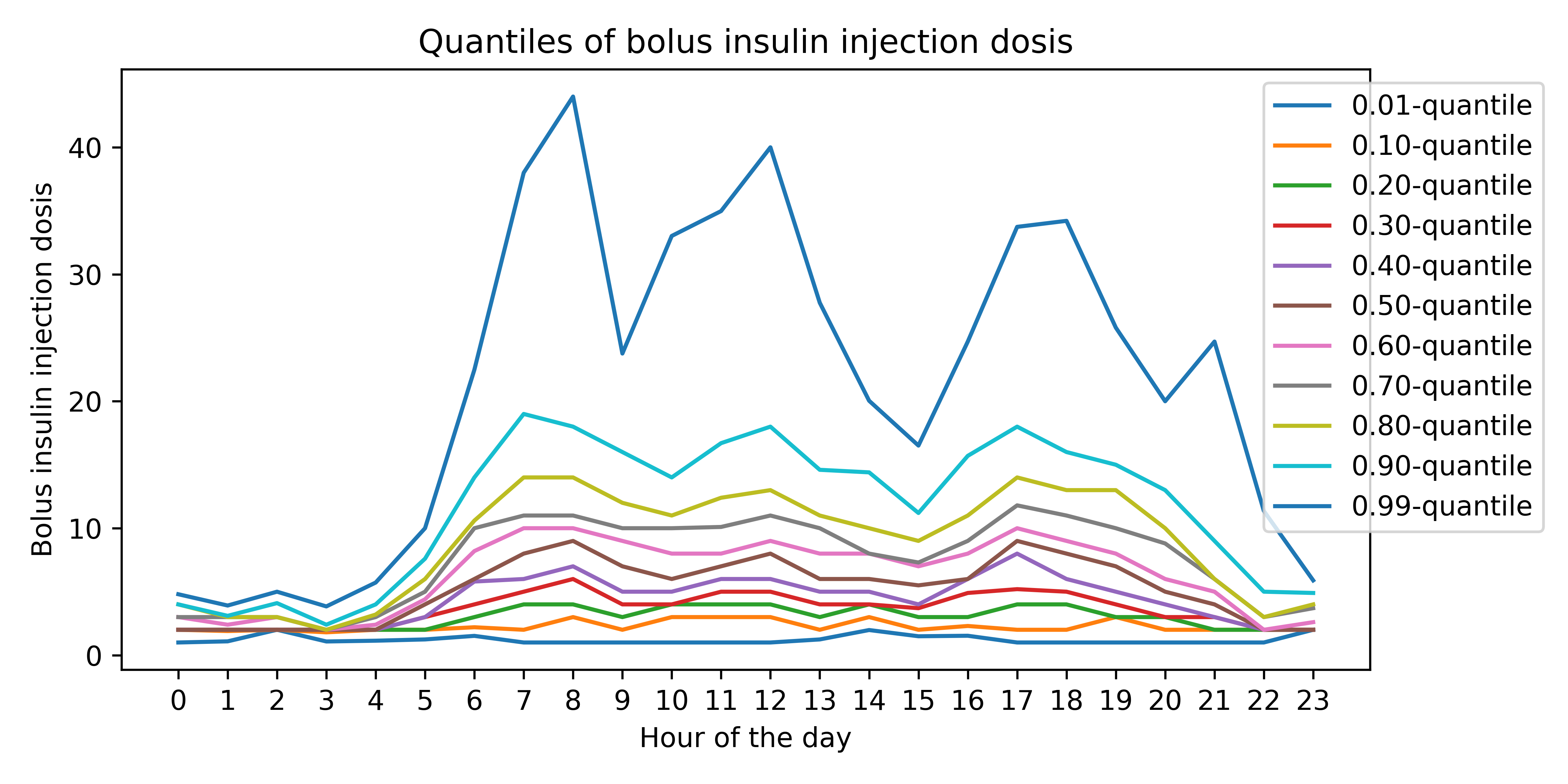}}
\end{figure*}



\begin{figure*}[htbp]
\centering
\floatconts
{fig:online}
{\caption{Probabilistic online prediction $p_{\phi}(\bs{y} \vert \bs{x}, \bs{c} )$ for different splits of past and future windows.}}
{\includegraphics[width=0.95\linewidth]{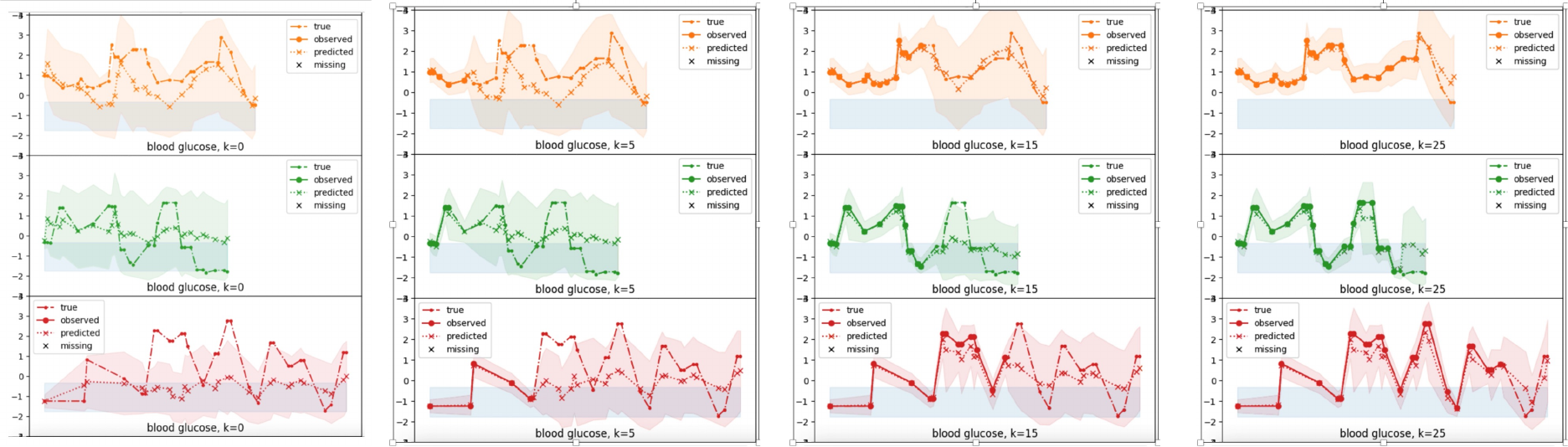}}
\end{figure*}

\begin{table*}[h]
    \centering
    \begin{tabular}{lccc}
        \toprule
        \textbf{Model} & \textbf{MAE} & \textbf{RMSE} \\
        \midrule
        Baseline: Past average glucose level of the patient & 2.671 & 3.671 \\
        Baseline: Average glucose level of all patients at this time & 3.049 & 3.927 \\
        Transformer (using glucose data only)  & 1.840 & 2.625 \\
        Transformer (using glucose, insulin, carbs data)  & 1.789 & 2.616 \\
        \bottomrule
    \end{tabular}
    \caption{Mean performance of the y-prediction network across 30 random 50/50 train-validation splits. Absolute Error (MAE) and Root Mean Squared Error (RMSE) for baseline and transformer models.}
    \label{tab:y_performance_model_comparison_table}
\end{table*}

\end{document}